\documentclass[10pt,twocolumn,letterpaper]{article}

\usepackage[pagenumbers]{cvpr} % To force page numbers, e.g. for an arXiv version

\usepackage{graphicx}
\usepackage{amsmath}
\usepackage{amssymb}
\usepackage{booktabs}
\usepackage[accsupp]{axessibility}  % Improves PDF readability for those with disabilities.

\usepackage[pagebackref,breaklinks,colorlinks]{hyperref}

\usepackage[capitalize]{cleveref}
\crefname{section}{Sec.}{Secs.}
\Crefname{section}{Section}{Sections}
\Crefname{table}{Table}{Tables}
\crefname{table}{Tab.}{Tabs.}

\def\cvprPaperID{13} % *** Enter the CVPR Paper ID here
\def\confName{CVPR}
\def\confYear{2023}

\begin{document}

\title{RB-Dust -- A Reference-based Dataset for Vision-based Dust Removal}

\author{Peter Buckel\textsuperscript{1,2}, Timo Oksanen\textsuperscript{2}, Thomas Dietmueller\textsuperscript{1}\\ \\
\textsuperscript{1}Baden-Wuerttemberg Cooperative State University (DHBW) Ravensburg, Germany\\
\textsuperscript{2}Technical University of Munich, Germany; Chair of Agrimechatronics;\\
 Munich Institute of Robotics and Machine Intelligence (MIRMI)\\
\url{www.agriscapes-dataset.com}} %

\maketitle

%%%%%%%%% ABSTRACT
\begin{abstract}
  Dust in the agricultural landscape is a significant challenge and influences, for example, the environmental perception of autonomous agricultural machines. Image enhancement algorithms can be used to reduce dust. However, these require dusty and dust-free images of the same environment for validation. In fact, to date, there is no dataset that we are aware of that addresses this issue. Therefore, we present the agriscapes RB-Dust dataset, which is named after its purpose of reference-based dust removal. It is not possible to take pictures from the cabin during tillage, as this would cause shifts in the images. Because of this, we built a setup from which it is possible to take images from a stationary position close to the passing tractor. The test setup was based on a half-sided gate through which the tractor could drive. The field tests were carried out on a farm in Bavaria, Germany, during tillage. During the field tests, other parameters such as soil moisture and wind speed were controlled, as these significantly affect dust development. We validated our dataset with contrast enhancement and image dehazing algorithms and analyzed the generalizability from recordings from the moving tractor. Finally, we demonstrate the application of dust removal based on a high-level vision task, such as person classification. Our empirical study confirms the validity of RB-Dust for vision-based dust removal in agriculture. 
\end{abstract}

%%%%%%%%% BODY TEXT
\section{Introduction}
\begin{figure}[t]
  \centering
  \includegraphics[width=0.99\columnwidth]{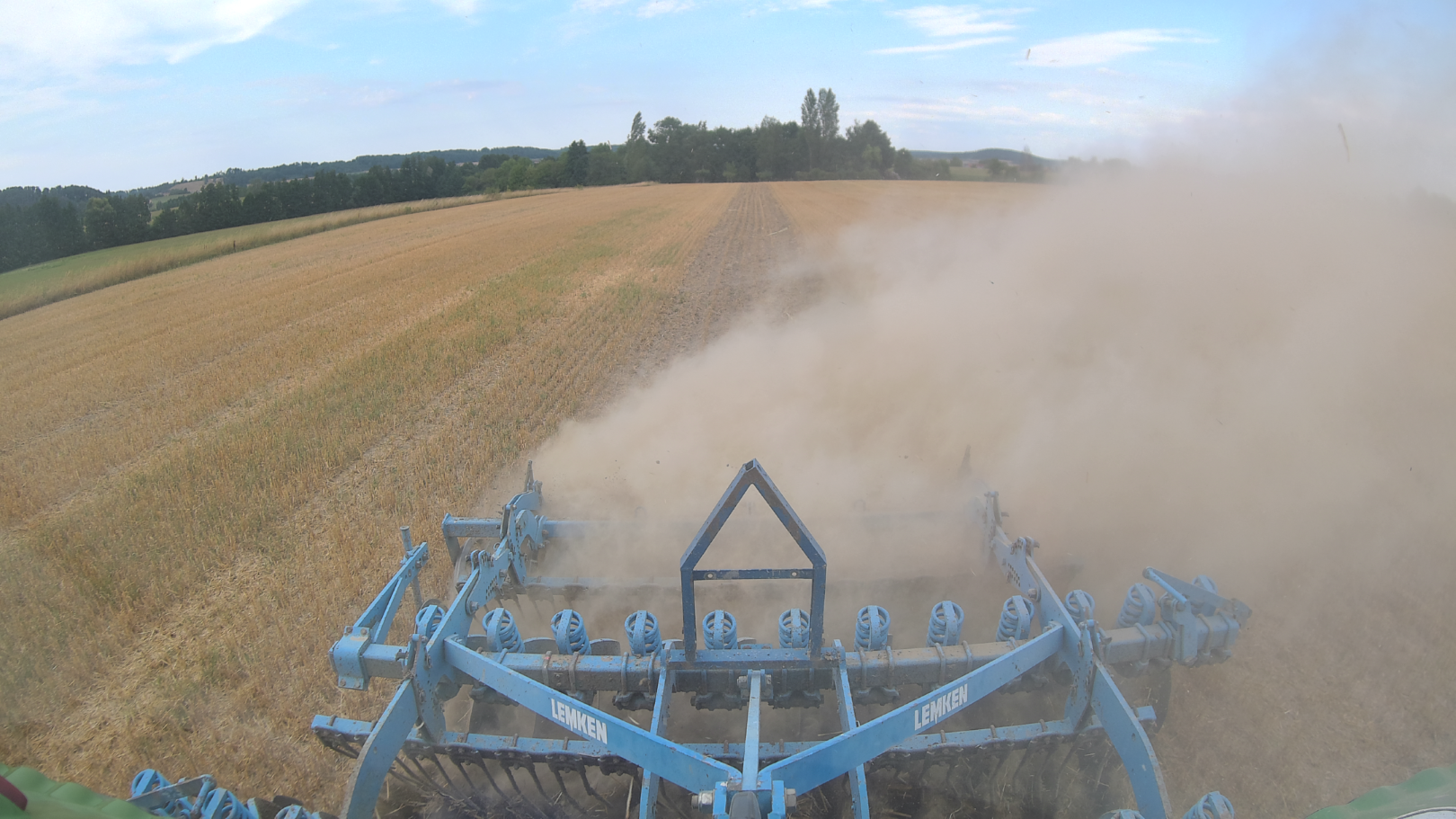}
  \caption{Dust raised during tillage.}
  \label{fig:dustcabin}
\end{figure}

Climate change is expected to cause longer periods of drought. 
These can lead to major problems, especially in agriculture. In particular, the effects on erosion, yield and human health have already been intensively researched~\cite{zhang2021characteristics,arslan2012particulate}. 
Dust resulting from drought also reduces the driver's visibility (see Figure \ref{fig:dustcabin}). 
With increasing automation, sensors are more frequently used, including cameras. Dust often covers a large part of images captured by cameras, and this can have a significant impact on the environmental perception of autonomous driving functions, such as the recognition of persons.
\newline
There are comparable problems in other areas. In particular, the topic of haze removal has been extensively researched. Consequently, methods have been developed which solve Koschmieders law~\cite{koschmieder1924theorie} for haze removal. To do this, different priors based on hazy and haze-free images are defined~\cite{fattal2014dehazing,he2010single,ancuti2010fast,ancuti2013single,tan2008visibility,fattal2008single}. Existing approaches could prove that the removal of haze has a significant influence on the detection of vehicles~\cite{li2017aod}. 
\newline
Although dust and haze have different physical properties, both scatter light. Thus, the atmospheric scattering model can also be used to analyze dust. There is no reference-based dataset that allows the validation of dust removal algorithms. 
A dataset must fulfil several requirements. The most important of which is that for every image with dust, there needs to be a dust-free equivalent from the same position. Normally, the camera is mounted on the tractor, or in the tractor cabin. However, this makes it difficult to take pictures from the same position with and without dust during tillage. 
\newline
Another approach would be to place the camera on a tripod and drive the tractor past the camera. This results in the required images, but the dust distribution in the images greatly differs from images out of the cabin. Therefore, we present a new approach in this paper.
\newline
\newline
The major contributions of our paper are:
\begin{itemize}
  \item We developed a measuring setup for flexible recording of dust-free and dusty images. The setup is based on a stationary, half-sided gate which is attached to a front loader.
  \item We created RB-Dust, a new dataset for validating dust removal methods. The dataset contains 200 images from three fields. It is publicly available.
  \item Furthermore, we verified the proposed setup with dehazing and contrast enhancement algorithms.
  \item We also demonstrate that the results of the methods are generalizable from images from the tractor cabin.
  \item Finally, we show how high-level vision tasks such as person detection can be improved by dust reduction.
\end{itemize}\section{Related Work}
Image enhancement algorithms require datasets with and without noise for development and validation. In our case the noise is dust. We are not aware of any dataset for dust removal in agriculture. A similar area of research is dehazing.
Therefore, we analyzed the datasets in the field of image dehazing to define the requirements and the structure of the new dataset. In addition, we presented the metrics used to evaluate image quality.
\subsection{Image Dehazing Datasets}
Dehazing is a popular field in image enhancement, with various different datasets.
In general, a distinction must be drawn between realistic and synthetic data. Realistic data is recorded outside under real environmental conditions. The haze is either real or created with a haze machine~\cite{ancuti2020nh}. Another way is to create artificial haze in images. The most current synthetic haze methods use the atmospheric scattering model and a corresponding depth map to synthesize haze~\cite{fattal2014dehazing}.
Table \ref{hazedatasets} provides an overview of the different datasets.
\newline
Li et al.~\cite{li2018benchmarking} introduced a synthetic, real-world dataset, called REalistic Single Image DEhazing (RESIDE), for subjective and objective evaluation. The dataset is based on the indoor NYU2~\cite{Silberman:ECCV12} and Middlebury stereo~\cite{scharstein2003high} dataset. The training set consists of 13,990 hazy images, which are synthesized using 1,399 haze-free images. The test data set consists of two parts. First is the synthetic objective testing set (SOTS), which contains 500 synthetic indoor hazy images. Second, there is the hybrid subjective testing set (HSTS) with ten synthetic and ten real-world outdoor images. Furthermore, the authors introduced RESIDE-${\beta}$ for validation of high-level vision tasks, such as object detection. Thus, they generated the synthetic outdoor training set (OTS) with 72,135 images and the real-world task-driven testing set (RTTS) with 4,322 images. In RTTS, cars, bicycles, motorbikes, persons, and buses are annotated~\cite{li2018benchmarking}. 
\newline
The non-homogenous realistic haze (NH-HAZE) dataset~\cite{ancuti2020nh} contains 55 outdoor image pairs. This dataset focuses on the fact that haze is not always homogeneously distributed. For this purpose, non-homogeneous haze was imitated with the help of a professional generator.
\newline
Zhao et al.~\cite{zhao2020dehazing} presented the benchmark dataset for dehazing evaluation (BeDDE) for full reference image quality assessment. The BeDDE contains 208 images from 23 Chinese cities, with three levels of haze: light, medium, and heavy.
\newline
The Dense-Haze~\cite{ancuti2019dense} dataset focuses on outdoor scenes, with haze generated by haze machines. It contains 33 image pairs of various scenes. In comparison with other datasets, Dense-Haze is more challenging due to its density.
\newline
Ancuti et al.~\cite{ancuti2018ohaze} presented an outdoor-based haze dataset called O-HAZE. The images are acquired with the same illumination with real haze and haze generated by haze machines. In total, it includes 45 different outdoor scenarios.
\newline
In contrast, the same authors also introduced I-HAZE~\cite{ancuti2018ihaze}, an indoor dehazing dataset with 35 images. In addition, a MacBeth color checker was placed in each shot. This allows color calibration and therefore improvement of the algorithms.
\newline
D-HAZY~\cite{ancuti2016d} contains over 1,400+ image pairs and is built upon Middlebury~\cite{scharstein2003high} and NYU2~\cite{Silberman:ECCV12}. It synthesizes haze based on the corresponding depth map from the datasets. Consequently, it was possible to create a dataset approaching reality~\cite{ancuti2016d}.
\newline
Zhang et al.~\cite{zhang2017hazerd} proposed HazeRD, a synthesized haze dataset, based on 15 outdoor images.
\newline
The foggy road image database~\cite{tarel2010improved} and FRIDA2~\cite{tarel2012vision} contain 90 images from 18 road scenes and 330 images from 66 scenes, respectively.
\begin{table}[h]
  \resizebox{\columnwidth}{!}{%\linewidth \columnwidth
  \centering
  \begin{tabular}{lcccc}
  \hline
              & Real/Synthetic  & Indoor/Outdoor & Annotations & Number of Images \\ \hline
  RESIDE~\cite{li2018benchmarking}      & Synthetic       & Indoor         & No          & 14,510            \\
  RESIDE-${\beta}$~\cite{li2018benchmarking} & Real\&Synthetic & Outdoor        & Yes         & 76,457            \\
  NH-HAZE~\cite{ancuti2020nh}     & Real            & Outdoor        & No          & 55               \\
  BeDDE~\cite{zhao2020dehazing}       & Real            & Outdoor        & No          & 20               \\
  Dense-Haze~\cite{ancuti2019dense}  & Real            & Outdoor        & No          & 33               \\
  O-HAZE~\cite{ancuti2018ohaze}      & Real            & Outdoor        & No          & 45               \\
  I-HAZE~\cite{ancuti2018ihaze}      & Real            & Indoor         & No          & 35               \\
  D-HAZY~\cite{ancuti2016d}      & Synthetic       & Indoor         & No          & 1,400+            \\
  HazeRD~\cite{zhang2017hazerd}      & Synthetic       & Outdoor        & No          & 15               \\
  FRIDA~\cite{tarel2010improved}       & Synthetic       & Outdoor        & No          & 90               \\
  FRIDA2~\cite{tarel2012vision}      & Synthetic       & Outdoor        & No          & 330              \\ \hline
  \end{tabular}
  }
  \caption{Comparison of different image dehazing datasets.}\label{hazedatasets}
\end{table}
\subsection{Image Quality Assessment}
Before it is possible to evaluate the algorithm, it is necessary to understand what the goal of the evaluation is and what metrics are going to be used.
The metrics for validation can be distinguished between reference-based and reference-less quality assessment. The reference-less, also called blind image quality assessment metrics, are specialized metrics for specific-use cases~\cite{wang2004image}, such as the Fog-Aware Density Evaluator~\cite{choi2015referenceless}, which predicts the visibility of foggy images. The natural image quality evaluator (NIQE) uses a natural scene statistic model based on natural, undistorted images. Results show that NIQE could compete with other reference-based metrics~\cite{mittal2012making}. The biggest advantage here is that one input image is required for evaluation. However, these metrics are mostly specialized to specific-use cases.
\newline
The other metric is reference-based, e.g., dust-free and dusty images, for the evaluation of image quality. The most common metrics are the peak signal-to-noise ratio (PSNR) and the structural similarity index (SSIM). The PSNR calculates the mean square error (MSE) of pixels~\cite{peaksnr}: 
\begin{equation}
  PSNR=10\log_{10}(\frac{peakval^{2}}{MSE})
  \label{eq:PSNR}
\end{equation}
The SSIM can be calculated using the following formula~\cite{wang2004image}:
\begin{equation}
  SSIM(x,y)=\left[l(x,y)  \right]^{a}\cdot \left[c(x,y)  \right]^{\beta}\cdot \left[s(x,y)  \right]^{\gamma}
  \label{eq:SSIM}
\end{equation}
The SSIM computes the luminance ${l(x,y)}$, contrast ${c(x,y)}$, and structural information ${s(x,y)}$ of an image.
Ideally, with two equal images PSNR is very high and SSIM is equal to one.
Due to the fact that reference-less metrics alone may not provide information about the quality of the algorithms, NIQE, SSIM, and PSNR will be used in this work.
\section{RB-Dust Dataset}
This section introduces the RB-Dust dataset. For this purpose, we first collect and define requirements for the test setup. Then, the measurement setup and the measurement series, including the environmental conditions, are presented.
\subsection{Test Setup Requirements}
In this work, we used the reference-based SSIM and PSNR, which required images of the same scenery with and without dust. 
This is a major challenge.
The obvious solution would be to install a camera in the cabin and drive slowly, so no dust is raised, and then fast. However, it is almost impossible to get two identical pictures while driving. In addition, the double tillage in the images changes the structure of the soil and driving over uneven ground leads to vibrations on the tractor. These are partially absorbed by the cabin but not completely; and therefore, the position of the camera always changes, even if minimally. In a pixel-to-pixel analysis, this can have a significant impact on the results. As a result, the camera must be mounted in a fixed position.
Furthermore, various environmental factors influence dust generation and distribution. On the one hand, it must be ensured that the ground is dry enough, so dust is raised during tillage. On the other hand, the wind must blow only slightly yet steadily during the tests.
Therefore, we can define the following requirements for the measurement setup and the associated field tests:
\begin{itemize}
  \item The camera must be attached in a fixed position and be insensitive to vibrations.
  \item The mounting position should be as close as possible to the tractor to increase the generalizability.
  \item The camera should be easy to assemble and disassemble, and easy to transport. This should make it possible to take as many pictures as possible from different positions.
  \item Tillage should be shallow, typically 5 -- 10 cm.
  \item A state-of-the-art camera that could also be used later in vehicles is to be used.
  \item Field testing should be conducted during an extended dry period.
  \item Wind, brightness, and soil moisture should be constant during a test series (dusty vs dust-free images).
\end{itemize}
Based on the defined requirements, we specified the measurement equipment and test cases.
\subsection{Test Setup and Test Cases}
\begin{figure}[h]
  \centering
  \includegraphics[width=0.99\columnwidth]{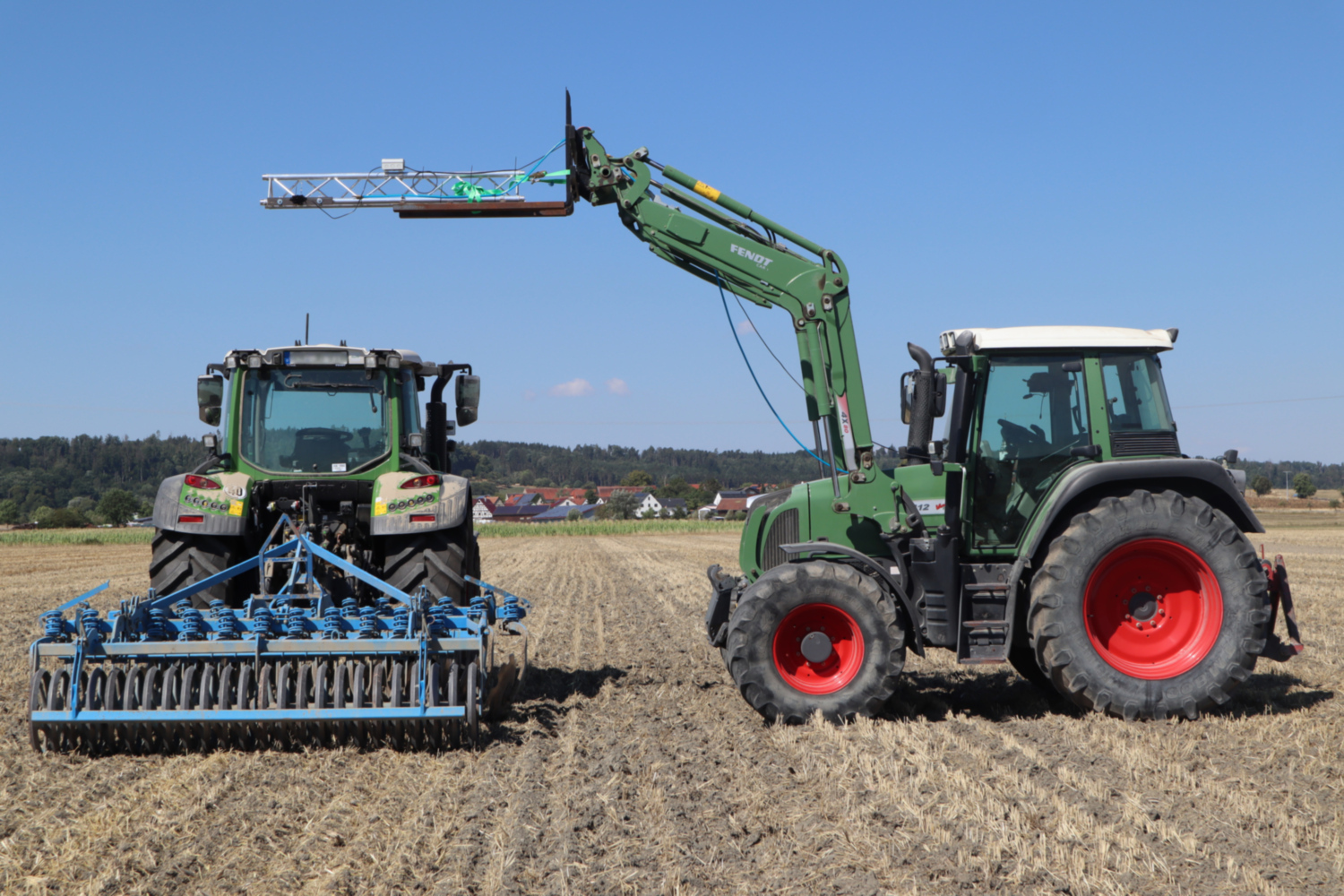}
  \caption{Dust measurement setup with a boom attached to a front loader, and a second tractor with a disc harrow.}
  \label{fig:setup}
\end{figure}
\begin{figure*}[t!]
  \centering
  \begin{subfigure}{0,99\linewidth}
    \includegraphics[width=0.245\columnwidth]{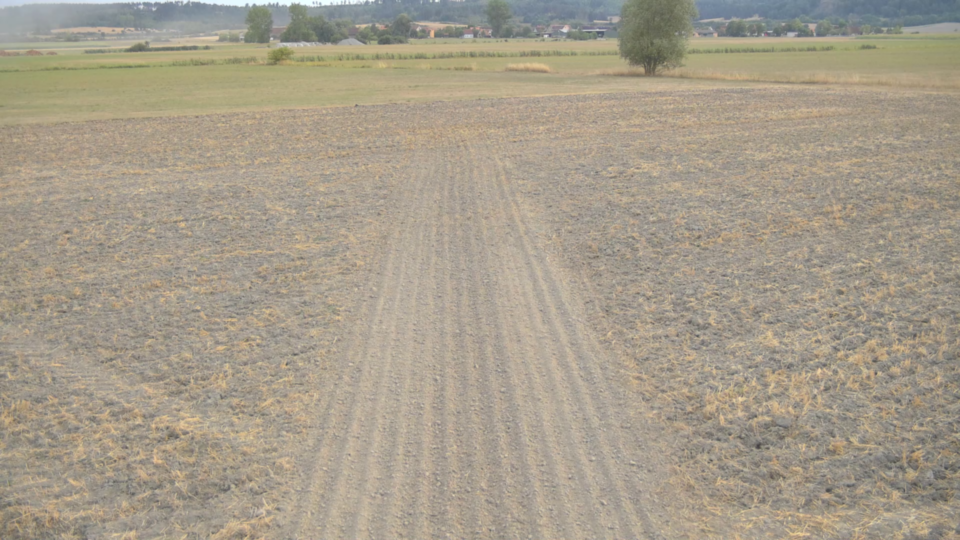}
    \hfill
    \includegraphics[width=0.245\columnwidth]{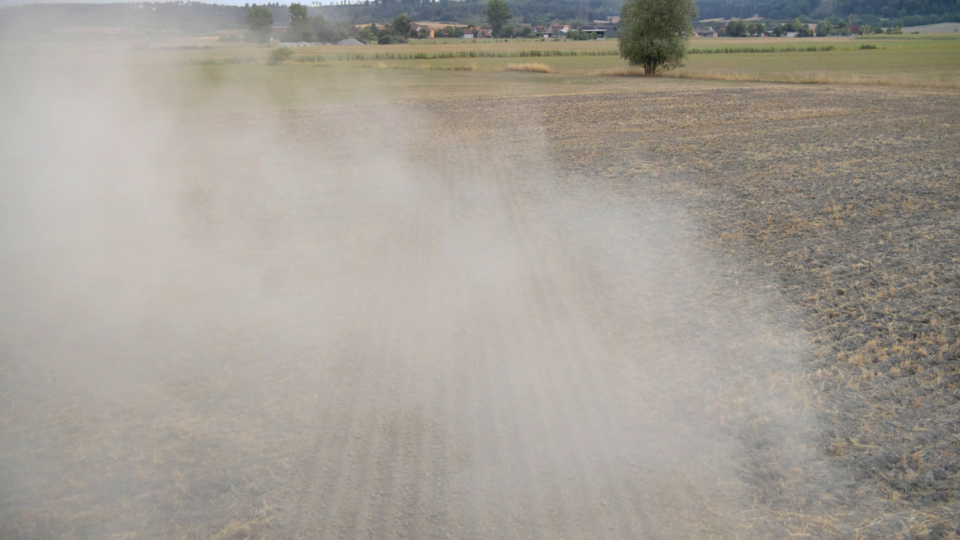}
    \hfill
    \includegraphics[width=0.245\columnwidth]{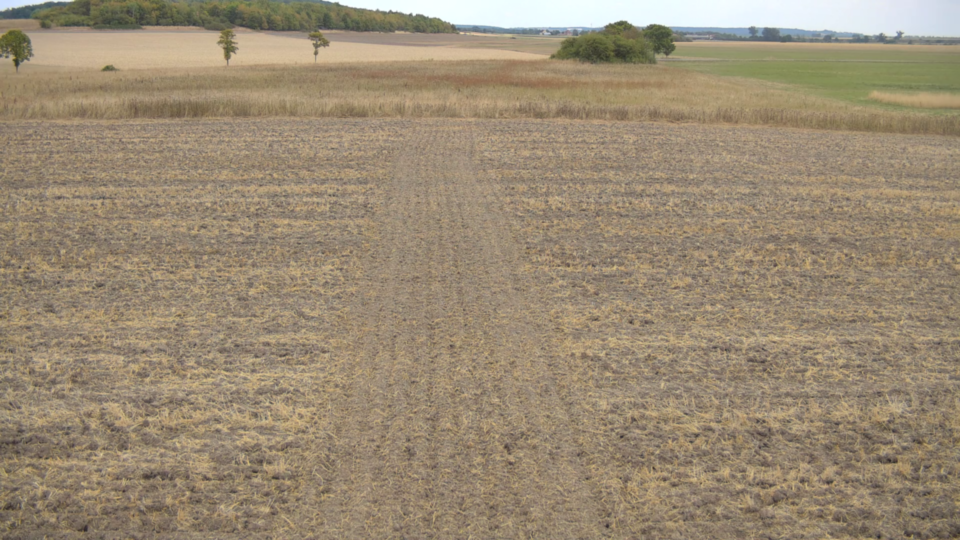}
    \hfill
    \includegraphics[width=0.245\columnwidth]{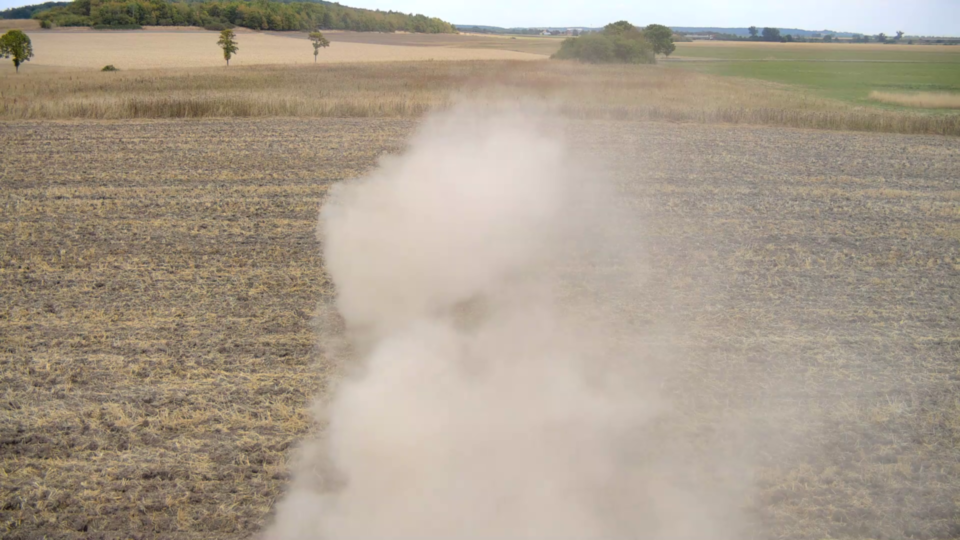}
    \qquad
    \includegraphics[width=0.245\columnwidth]{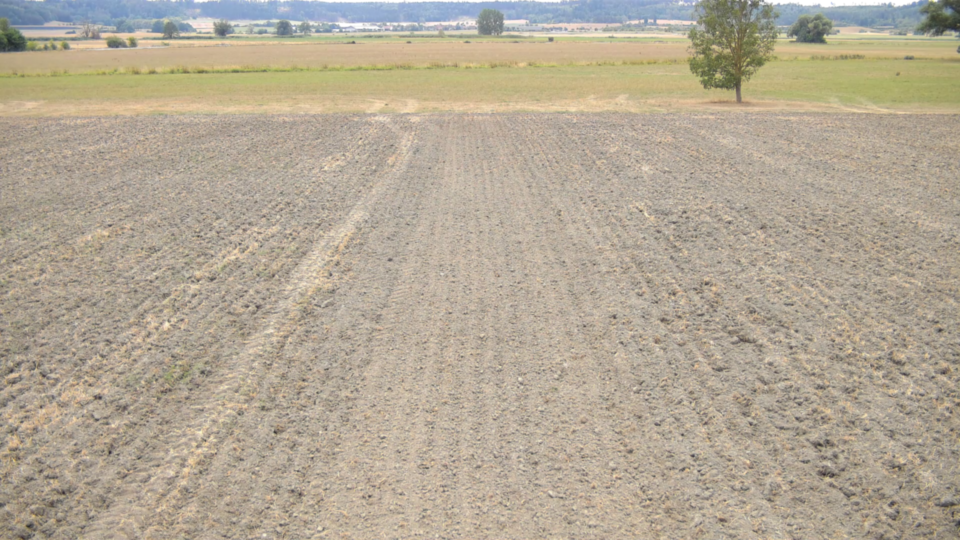}
    \hfill
    \includegraphics[width=0.245\columnwidth]{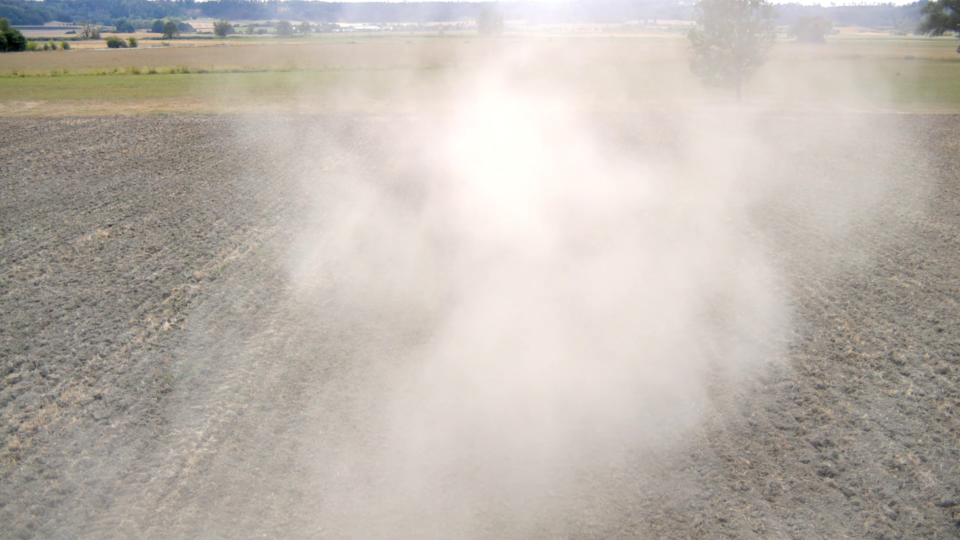}
    \hfill
    \includegraphics[width=0.245\columnwidth]{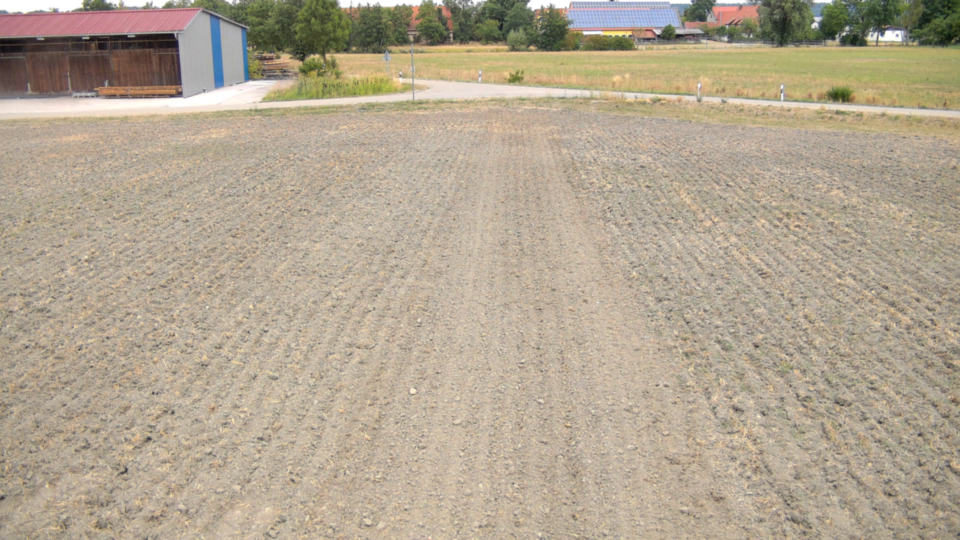}
    \hfill
    \includegraphics[width=0.245\columnwidth]{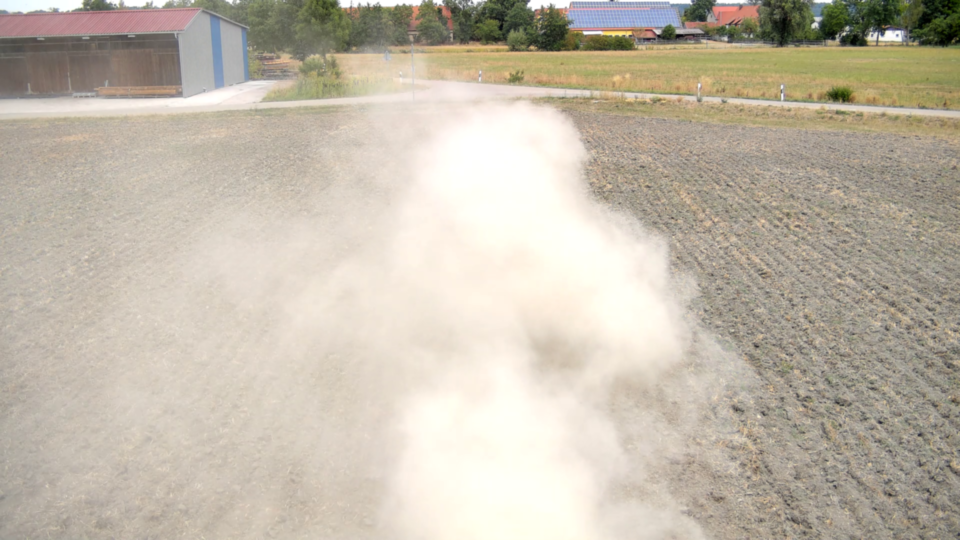}
    \qquad
    \subfloat[Dust-free GT]{\includegraphics[width=0.245\columnwidth]{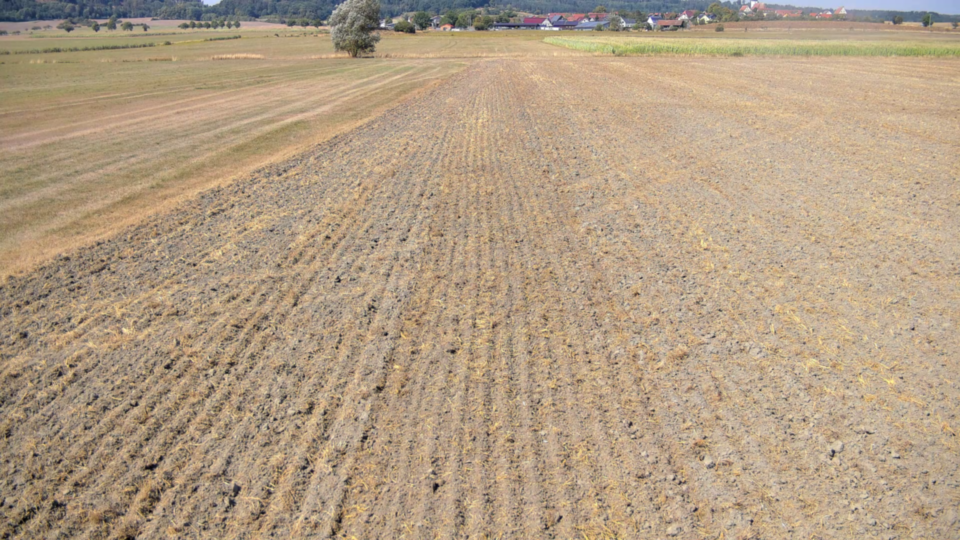}}
    \hfill
    \subfloat[Dusty view of (a)]{\includegraphics[width=0.245\columnwidth]{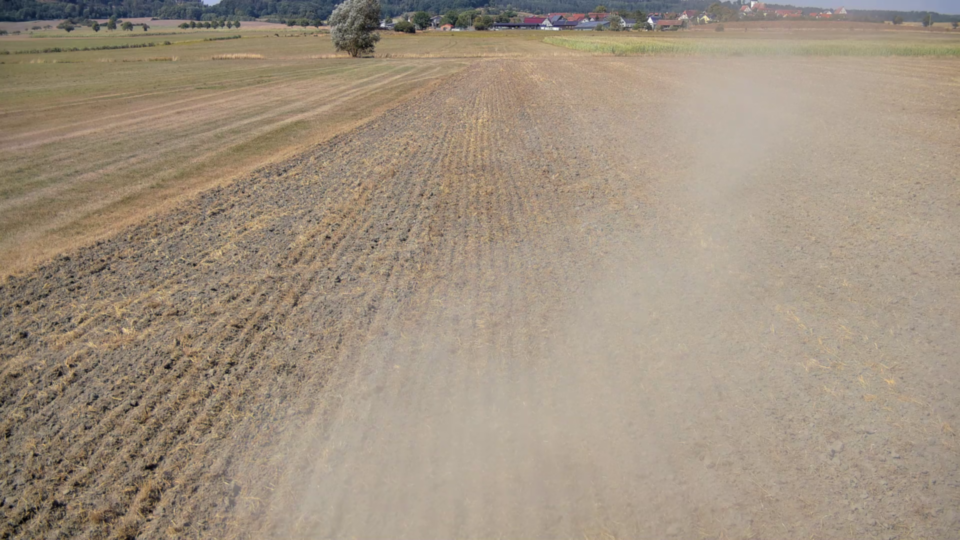}}
    \hfill
    \subfloat[Dust-free GT]{\includegraphics[width=0.245\columnwidth]{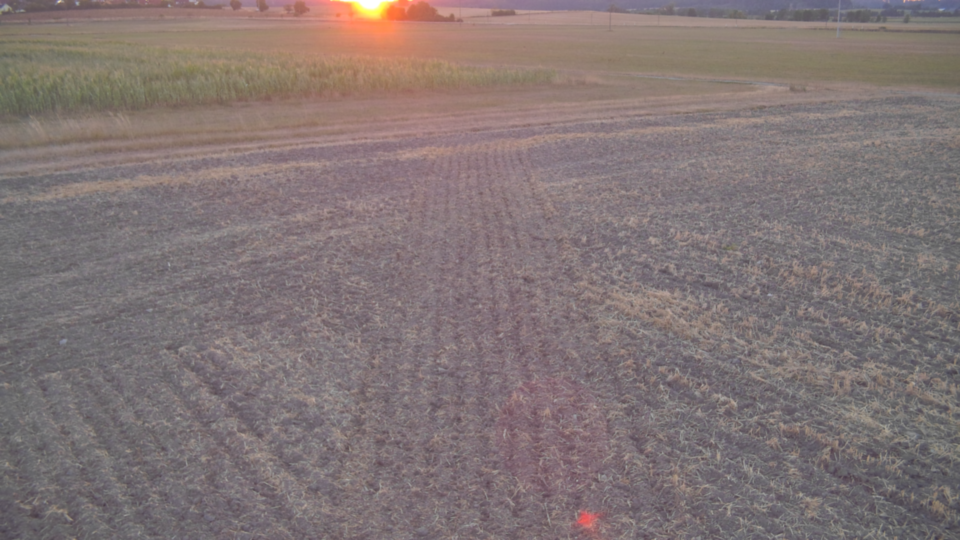}}
    \hfill
    \subfloat[Dusty view of (c)]{\includegraphics[width=0.245\columnwidth]{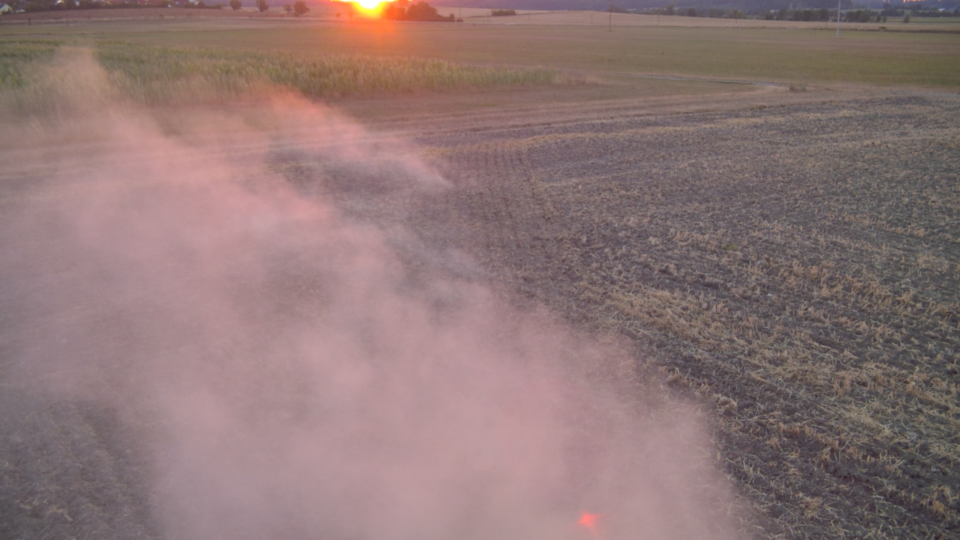}}
  \end{subfigure}
  \caption{Excerpt from the dataset with dusty images (b, d) and the corresponding dust-free images (a, c).}
  \label{fig:excerptdata}
\end{figure*}
The main idea of the proposed field test setup was to place the sensor as close as possible to the tractor but on a stationary object. The results are thereby presumably similar to those from a position on the tractor.
Therefore, we built a gate on which the sensors were mounted. This way, it was possible to record dusty and dust-free images while the tractor drove through. The test setup had to fulfil further requirements. In order for an agricultural vehicle to pass through, it had to have a height of at least 4 m and a boom-length long enough to safely drive through it. As a result, a tractor with a front loader was used, on which a boom was mounted. A tractor allowed easy transport and thus the possibility to record more data more easily.
The boom was based on an aluminum traverse. The traverse was mounted on a receptacle for the front loader forks. A compressed air hose (blue hose) was attached to the traverse to free the camera lens from dust. This saved time because it was not necessary to lower the front loader and clean the camera. The camera, in turn, was attached to the traverse. An HDR camera based on the AR0821 image sensor was used. The camera had a resolution of 8.1 MP, with a dynamic range of up to 140 dB, and was therefore well suited for challenging light conditions. The sensor had a size of ½ " and a pixel size of 2.1 \textmu m. For image capture, the camera used a rolling shutter. The lens had an 8 mm fixed focal length. The optical distortion of the lens wass less than 2.9\%. 
The captured and rectified images had a resolution of 1920 x 1080 pixels. This resulted in a field of view of 53.3° x 31.1° (horizontal x vertical). 
\newline
Camera parameters such as exposure or white-balancing were set manually.
The setup with the second tractor for tillage is shown in Figure \ref{fig:setup}.
The clearance height was set at approx. 4 m, and the boom had a length of approx. 3 m. The boom with the mounted camera was placed above the middle of the tillage track. The height and alignment of the camera was checked after each repositioning of the tractor.
In addition, environmental parameters were controlled during the field tests using a luxmeter, soil moisture sensor, and a weather station. This included temperature, air pressure, humidity, wind speed, soil moisture, and temperature. With the help of the sensors, it was ensured that the ambient conditions remained constant within a measurement series. Special attention was paid to constant light conditions, since these significantly influence light scattering by particles such as dust.
\newline
Various test cases were defined for the field tests.
In total, tests were carried out on three fields in Bavaria, Germany during August 2022, when it was very dry. This means that at a depth of 20 cm, no soil moisture was measurable. 
According to the international soil type triangle, the soil in the fields is classified as silt loam.
On each field, the measurement setup was placed at various positions with different backgrounds of the field. In addition, tests were carried out at different times of day, and thus under different lighting conditions. Furthermore, we ensured that the tests were performed only when the wind speed was less than 5 km/h. This ensured that the dust remained in the camera's field of view for as long as possible and did not immediately disappear.
For tillage, a disc harrow was used and tilled at approximately 15 km/h at a depth of approx. 5 cm. The variance of time of day, environment, and speed, and thus the amount of dust stirred up, ensured that the results would be generalizable and cover most of the standard scenarios in soil cultivation.
\subsection{Dataset Excerpt}
An excerpt from the dataset is shown in Figure \ref{fig:excerptdata}. It illustrates the dust-free ground truth (GT) data (a, c) and the corresponding dusty image (b, d) and shows that the images were recorded from the same position. Moreover, the dataset contains a variety of different backgrounds, such as electricity pylons, buildings, woods, hedges, and open areas. For the dataset, we made several measurements on the same fields. From each measurement, several images with different dust levels were randomly selected. The dataset contains 200 images and is available in PNG file format. 
\section{Verification of the Dataset}
We validated the previously presented dataset as follows:
First, we examined how resistant the measurement setup was to vibrations. 
After that, state-of-the-art enhancement and dehazing algorithms were presented and applied to the datasets. In the following, we present our verification process and discuss the generalizability from images taken from the tractor cabin.
\subsection{Verification of the Test Setup}
In the first step, we analyzed the static behavior of the test setup and thus the background noise. The passing tractor can lead to vibrations, and thus to vibration of the camera at the boom.  
For this purpose, we examined two dust-free images immediately after tillage. The time difference was 5s. 
The time shift was deliberately chosen to be very small, since the influence of changing light conditions, for example, should be kept low.
We conducted this experiment using ten different backgrounds and calculated the average SSIM. We obtained an SSIM of 0.9767. 
In an ideal case, the SSIM is 1, and thus the difference is 0.0233. This means that the camera on the boom was subject to minor fluctuations. 
This can have several causes:
\begin{itemize}
  \item Background noise: The images were recorded under real circumstances, where even low wind can lead to movements, of e.g., leaves or grass.
  \item Vibrations due to soil cultivation: The second tractor drove through the gate and cultivated the soil in the process. This occurred with a safety distance of approx. 1.5 m to the gate and thus to the second tractor. Nevertheless, it is possible that the vibrations were passed on to the camera via the tires, front loader, and boom.
  \item Dust in the air: The images were taken after each drive with the disc harrow through the gate. It is possible that even if the images look dust-free to the human eye, there were still dust particles in the air. This can lead to minor noise in the images.
\end{itemize}
These factors lead to deviations from the ideal value of one, especially in a pixel-to-pixel analysis, such as the one that the SSIM is based on. Nevertheless, the deviation is very small, and therefore it can be assumed that the images are suitable for further investigation. Also, it must be noted that these algorithms never achieve the ideal value.
\subsection{Baseline Experiments}
\begin{figure*}[t]
  \centering
  \begin{subfigure}{0.99\linewidth}
    \centering
    \includegraphics[width=0.122\columnwidth]{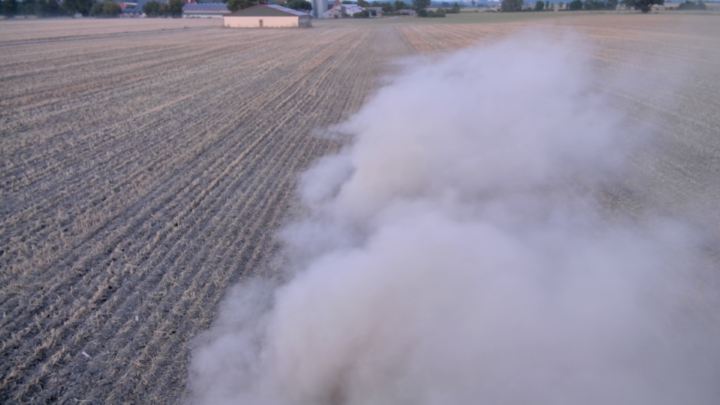}\hfill
    \includegraphics[width=0.122\columnwidth]{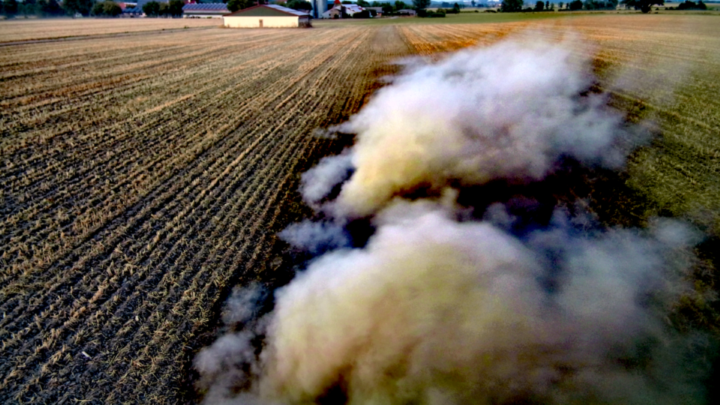}\hfill
    \includegraphics[width=0.122\columnwidth]{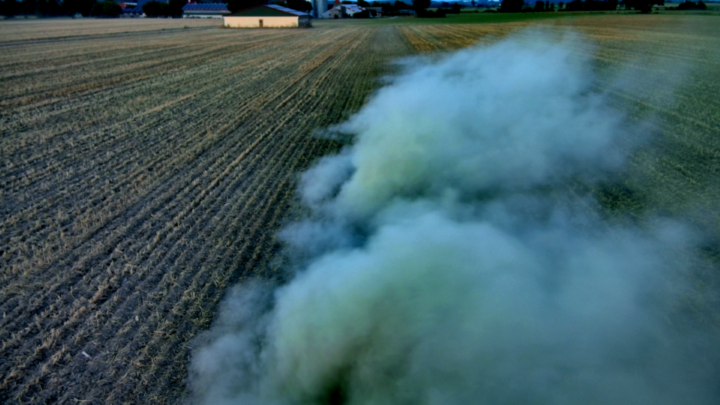}\hfill
    \includegraphics[width=0.122\columnwidth]{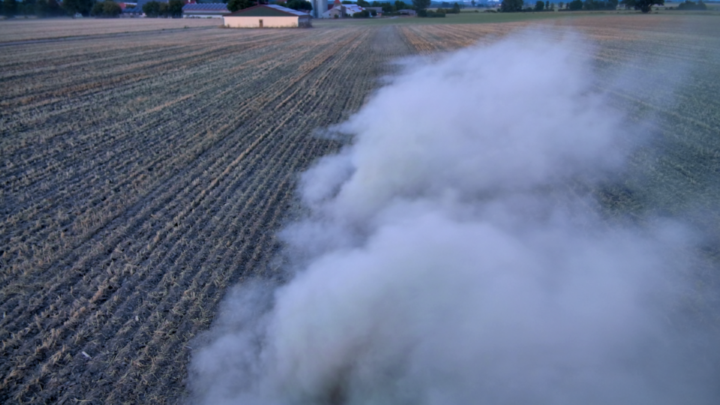}\hfill
    \includegraphics[width=0.122\columnwidth]{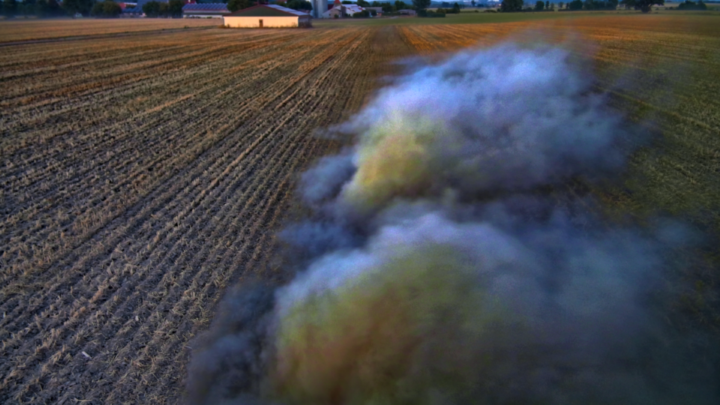}\hfill
    \includegraphics[width=0.122\columnwidth]{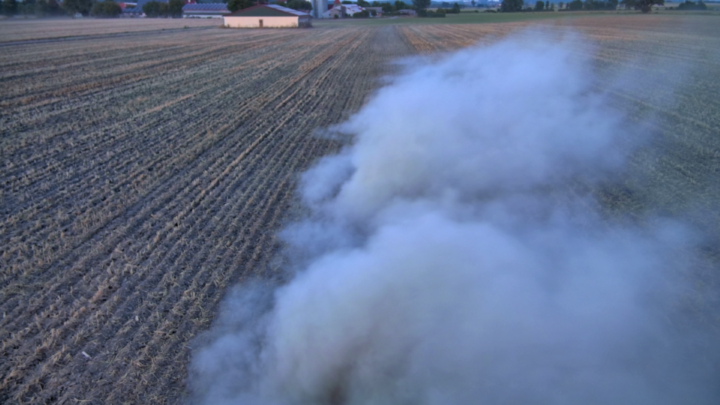}\hfill
    \includegraphics[width=0.122\columnwidth]{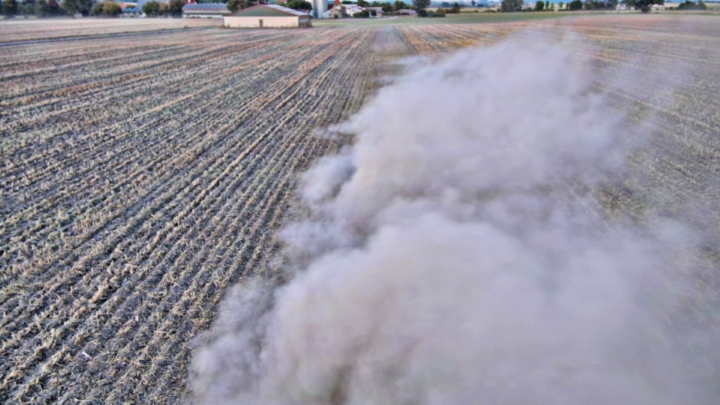}\hfill
    \includegraphics[width=0.122\columnwidth]{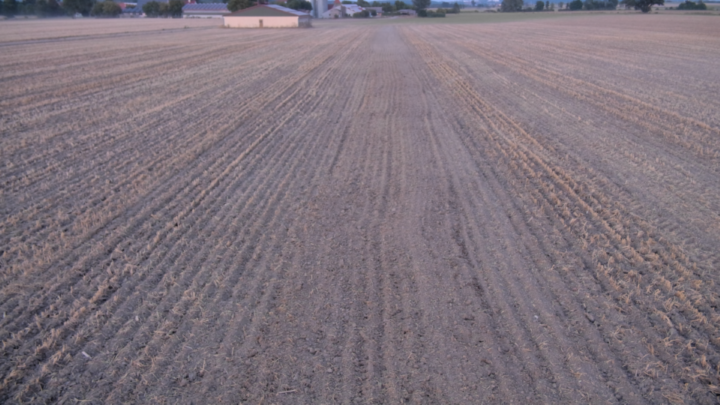}
    \qquad
    \includegraphics[width=0.122\columnwidth]{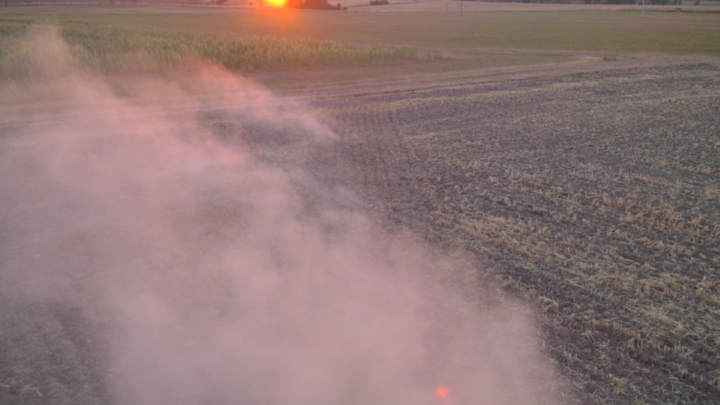}\hfill
    \includegraphics[width=0.122\columnwidth]{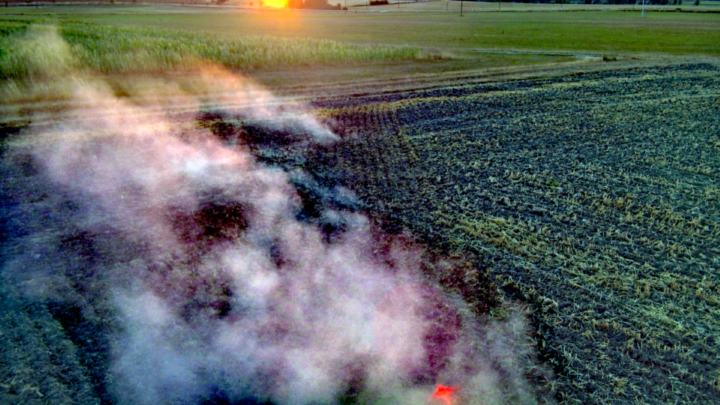}\hfill
    \includegraphics[width=0.122\columnwidth]{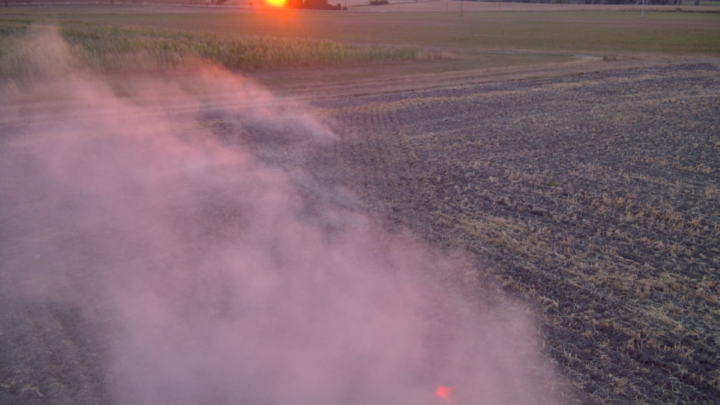}\hfill
    \includegraphics[width=0.122\columnwidth]{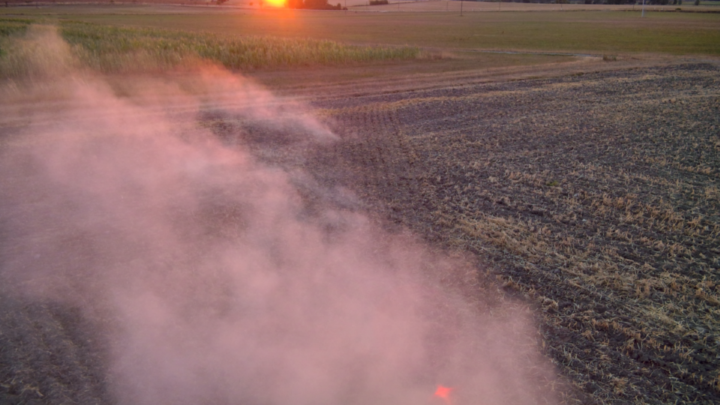}\hfill
    \includegraphics[width=0.122\columnwidth]{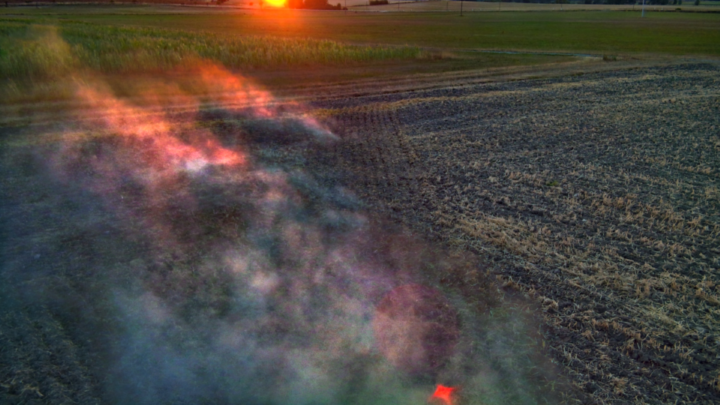}\hfill
    \includegraphics[width=0.122\columnwidth]{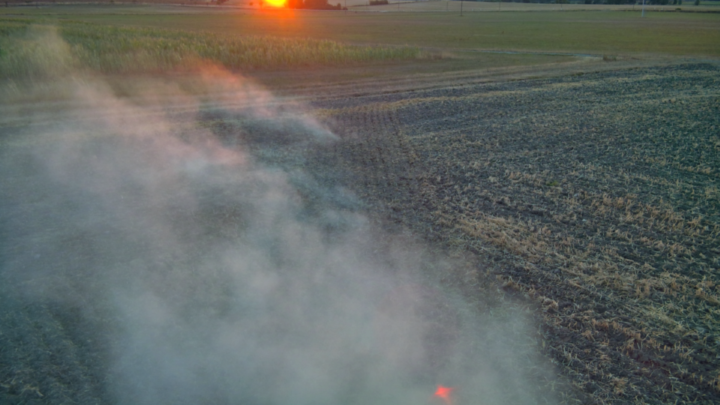}\hfill
    \includegraphics[width=0.122\columnwidth]{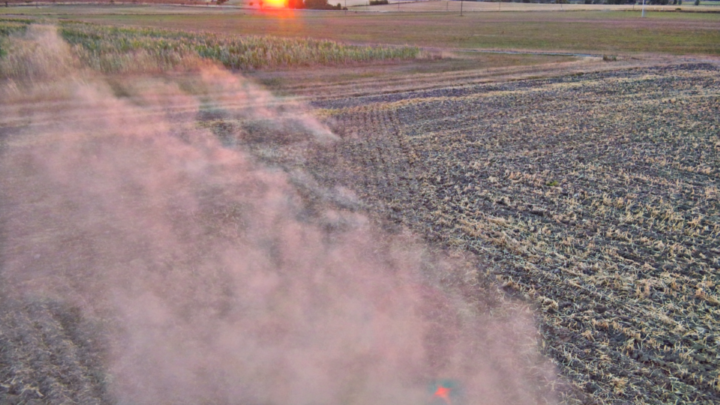}\hfill
    \includegraphics[width=0.122\columnwidth]{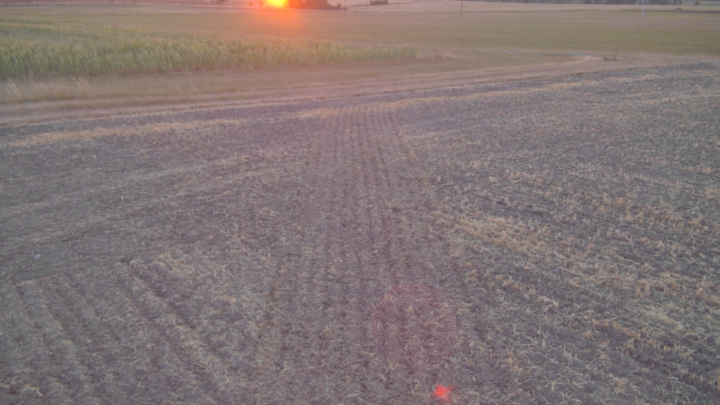}
    \qquad
    \includegraphics[width=0.122\columnwidth]{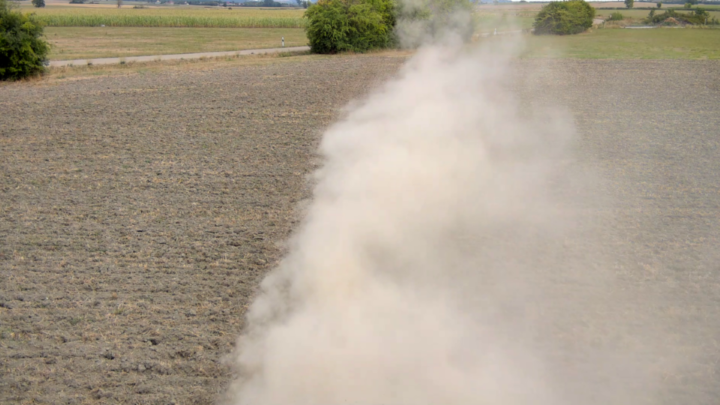}\hfill
    \includegraphics[width=0.122\columnwidth]{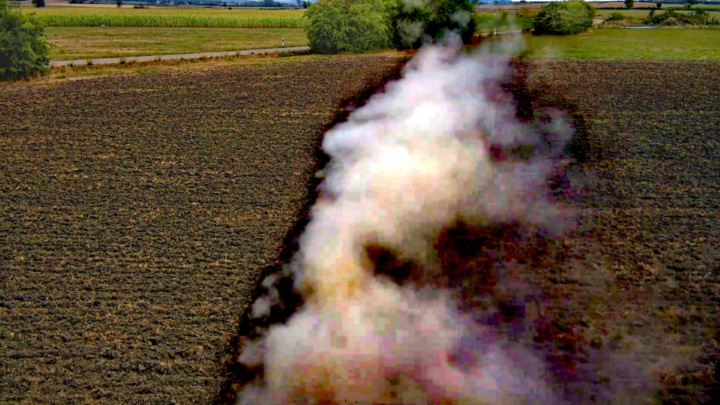}\hfill
    \includegraphics[width=0.122\columnwidth]{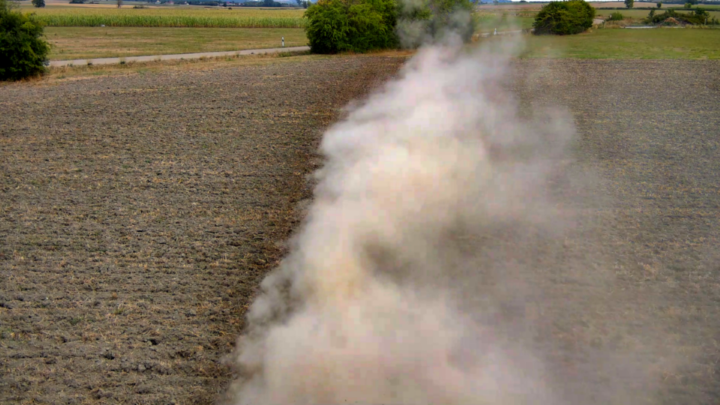}\hfill
    \includegraphics[width=0.122\columnwidth]{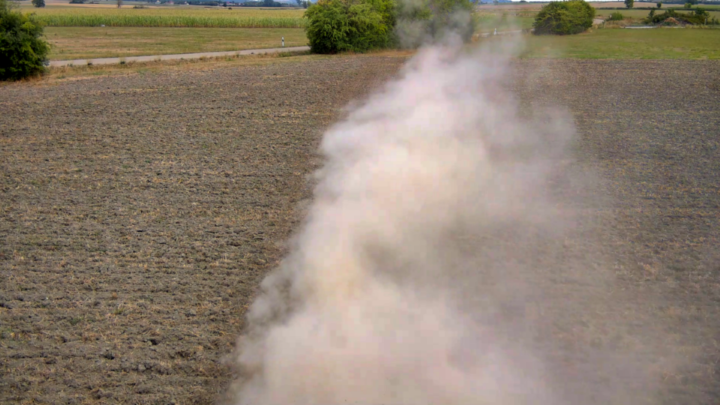}\hfill
    \includegraphics[width=0.122\columnwidth]{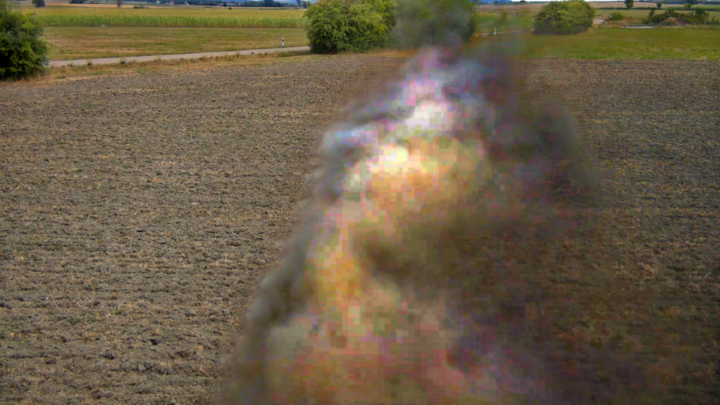}\hfill
    \includegraphics[width=0.122\columnwidth]{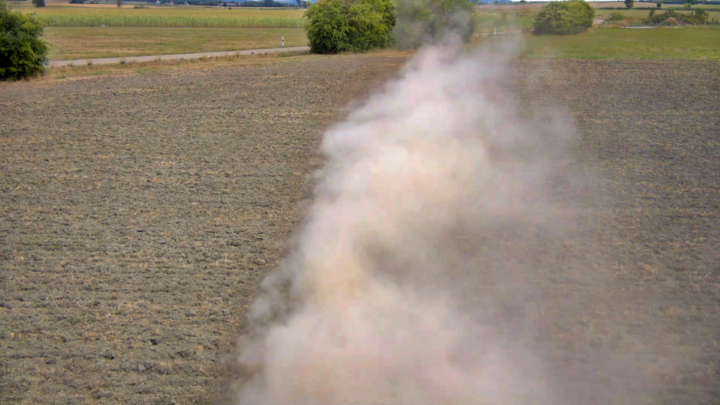}\hfill
    \includegraphics[width=0.122\columnwidth]{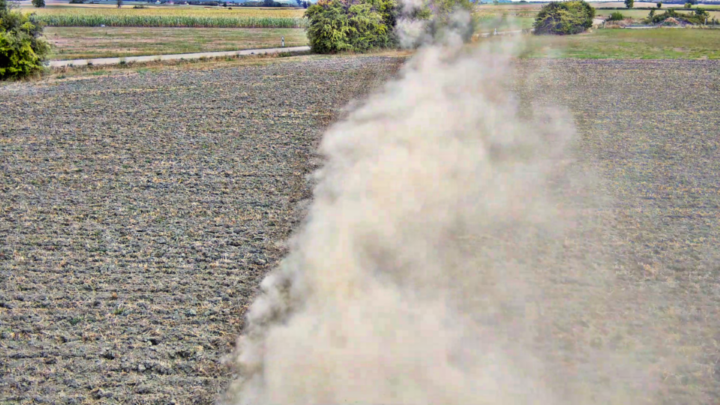}\hfill
    \includegraphics[width=0.122\columnwidth]{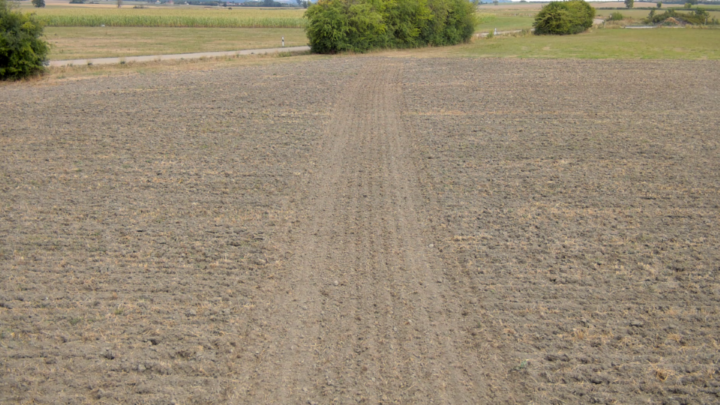}
    \qquad    
    \includegraphics[width=0.122\columnwidth]{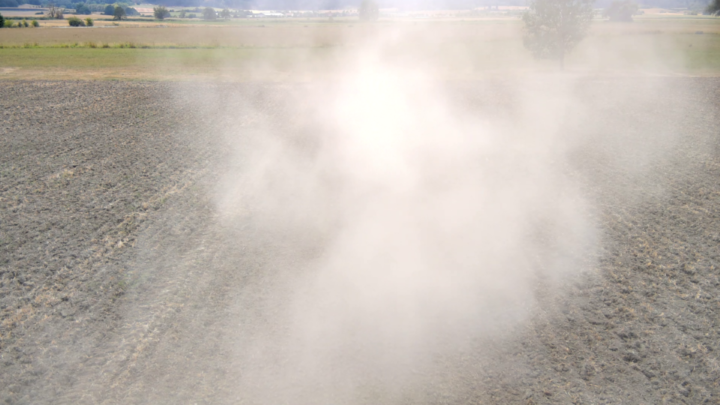}\hfill
    \includegraphics[width=0.122\columnwidth]{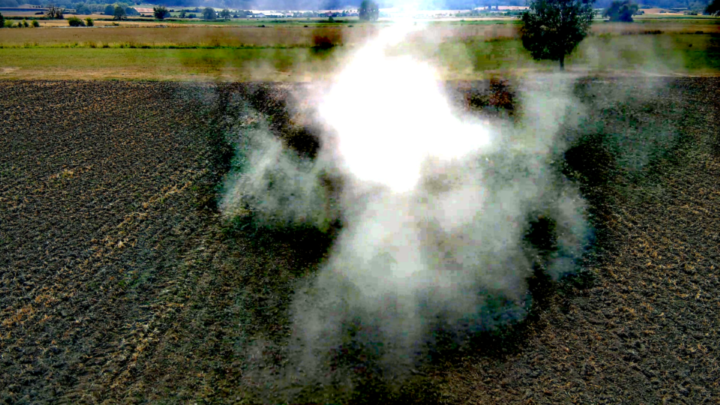}\hfill
    \includegraphics[width=0.122\columnwidth]{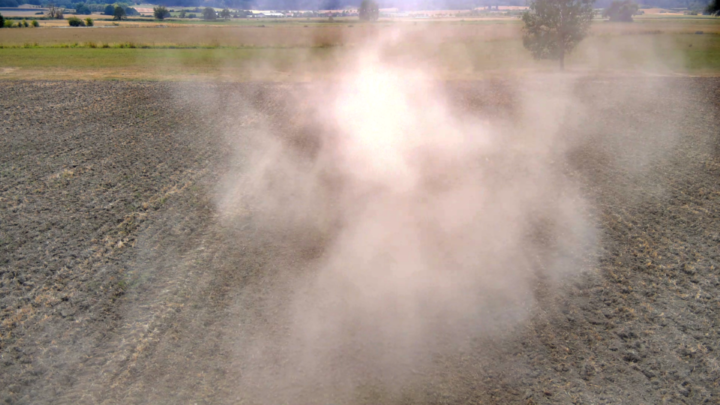}\hfill
    \includegraphics[width=0.122\columnwidth]{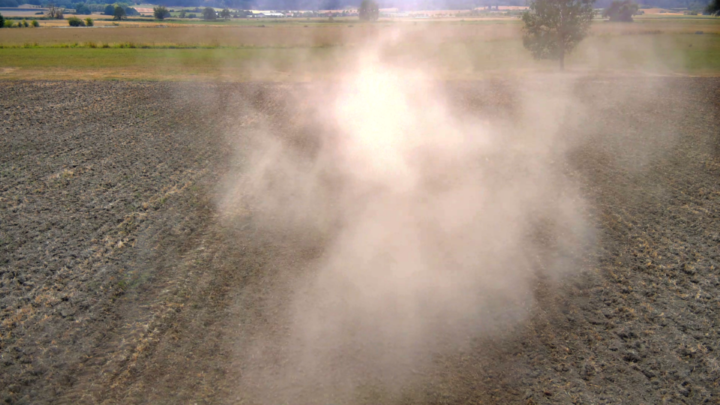}\hfill
    \includegraphics[width=0.122\columnwidth]{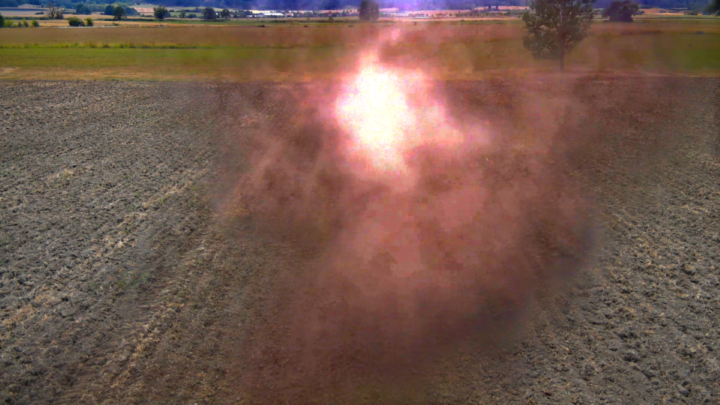}\hfill
    \includegraphics[width=0.122\columnwidth]{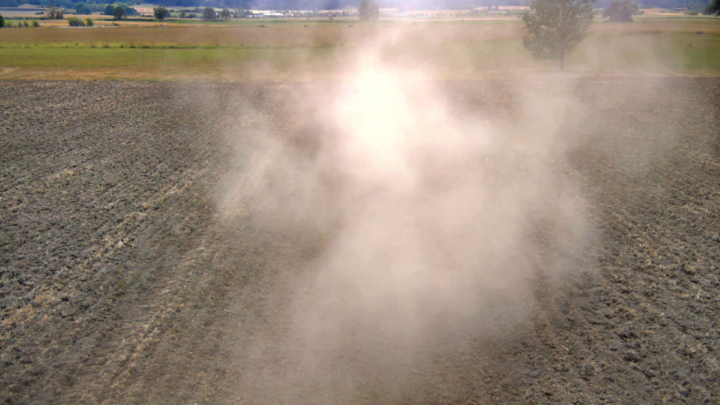}\hfill
    \includegraphics[width=0.122\columnwidth]{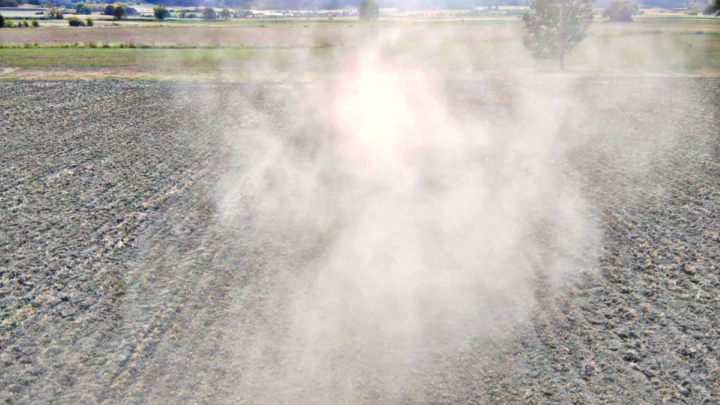}\hfill
    \includegraphics[width=0.122\columnwidth]{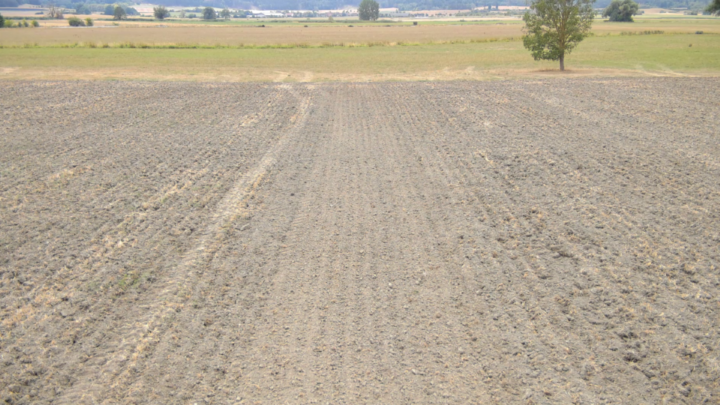}
    \qquad
    \includegraphics[width=0.122\columnwidth]{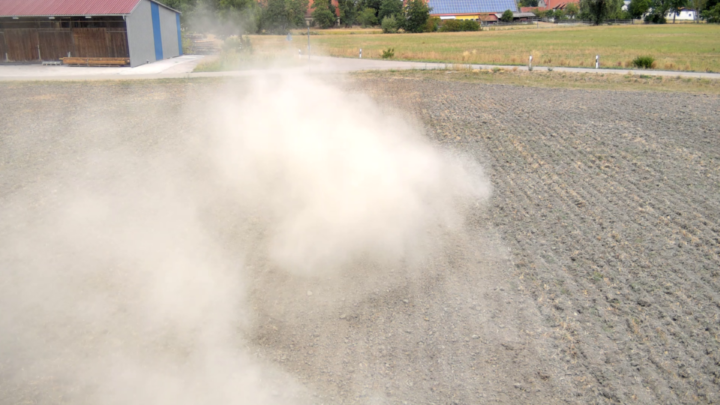}\hfill
    \includegraphics[width=0.122\columnwidth]{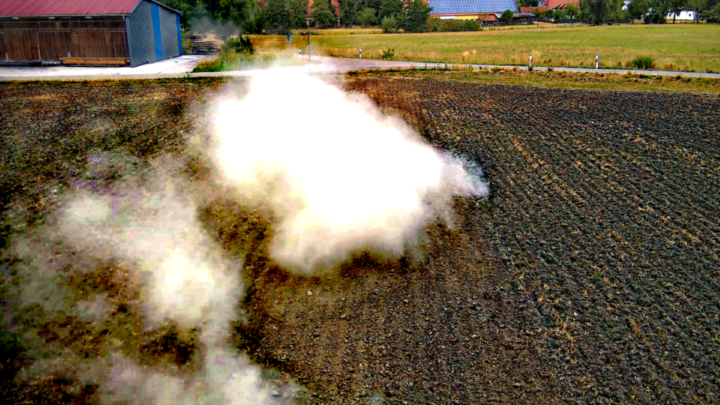}\hfill
    \includegraphics[width=0.122\columnwidth]{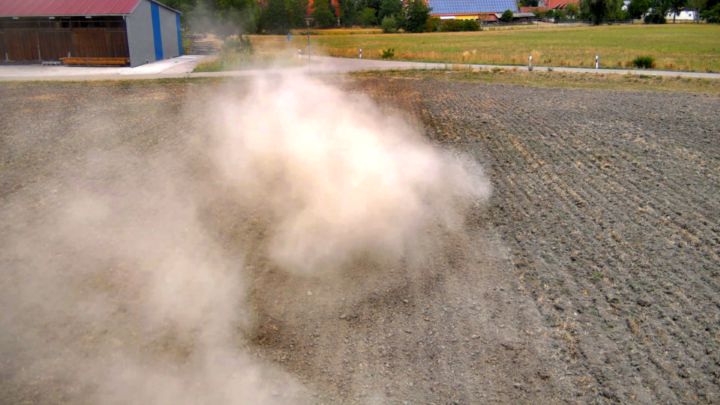}\hfill
    \includegraphics[width=0.122\columnwidth]{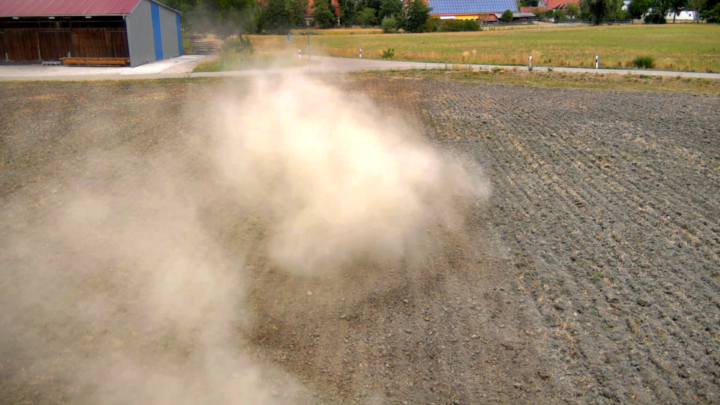}\hfill
    \includegraphics[width=0.122\columnwidth]{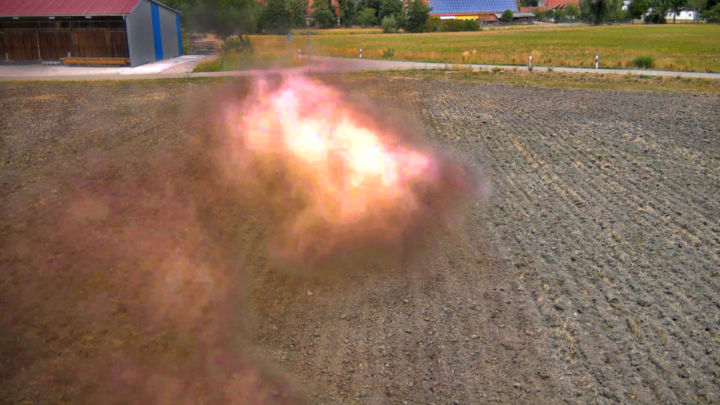}\hfill
    \includegraphics[width=0.122\columnwidth]{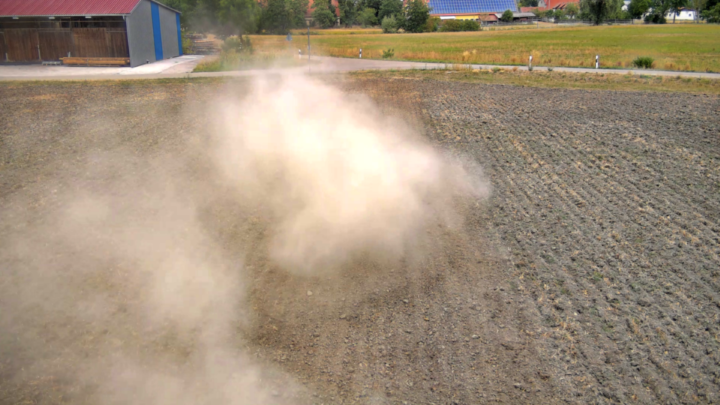}\hfill
    \includegraphics[width=0.122\columnwidth]{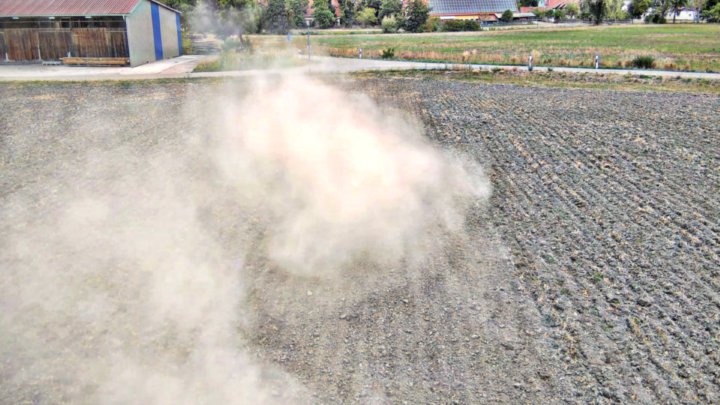}\hfill
    \includegraphics[width=0.122\columnwidth]{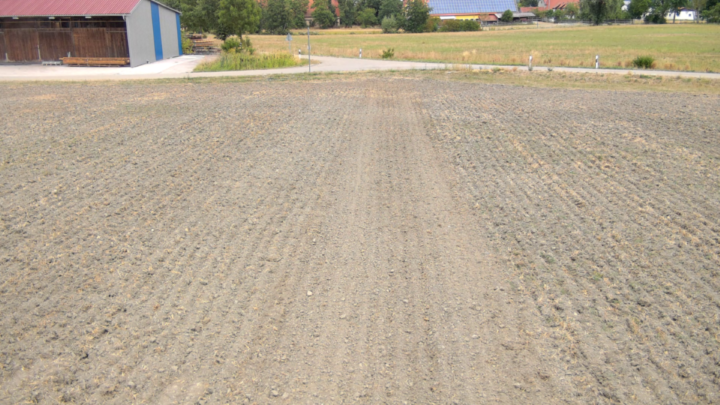}
    \qquad
    \subfloat[Dusty image]{\includegraphics[width=0.122\columnwidth]{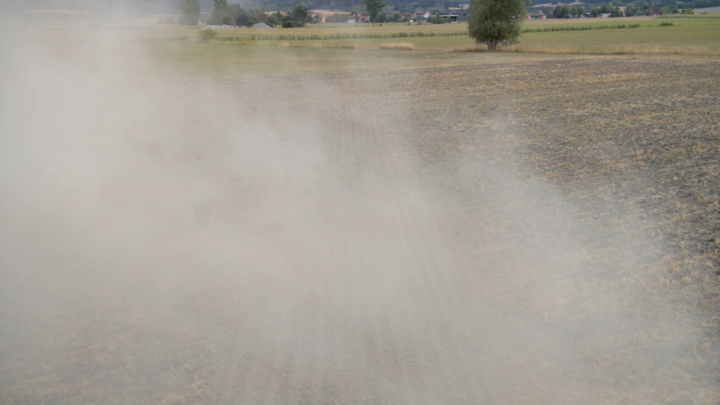}}\hfill
    \subfloat[NLD~\cite{berman2016non}]{\includegraphics[width=0.122\columnwidth]{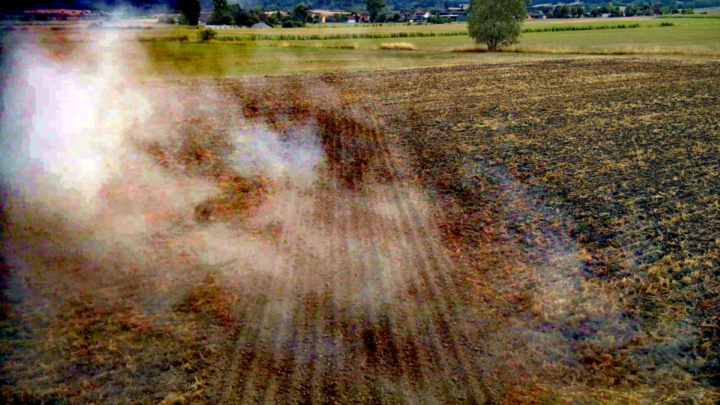}}\hfill
    \subfloat[MSCNN~\cite{ren2016single}]{\includegraphics[width=0.122\columnwidth]{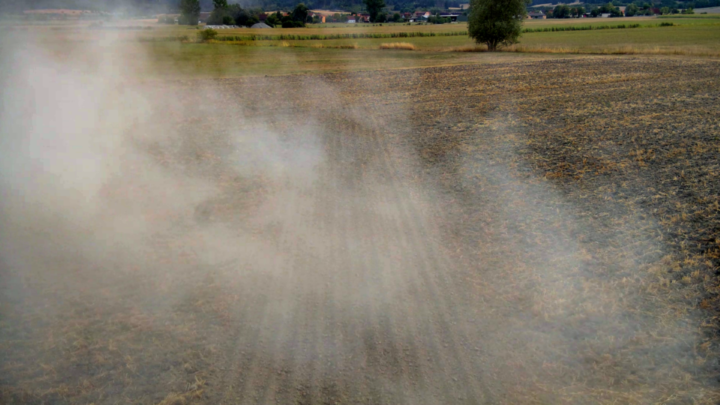}}\hfill
    \subfloat[\centering DehazeNet\newline\cite{cai2016dehazenet}]{\includegraphics[width=0.122\columnwidth]{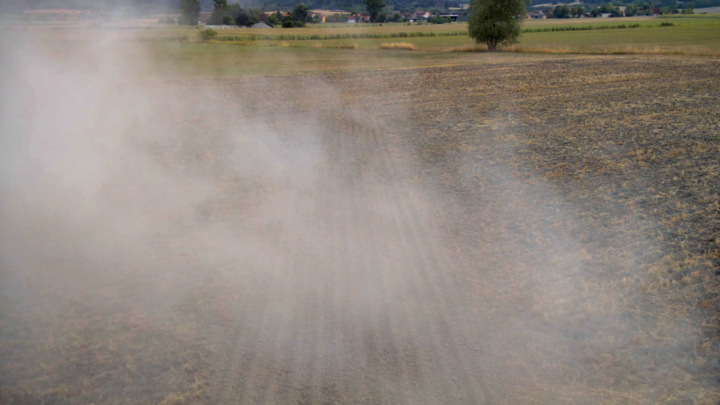}}\hfill
    \subfloat[DCP~\cite{he2010single}]{\includegraphics[width=0.122\columnwidth]{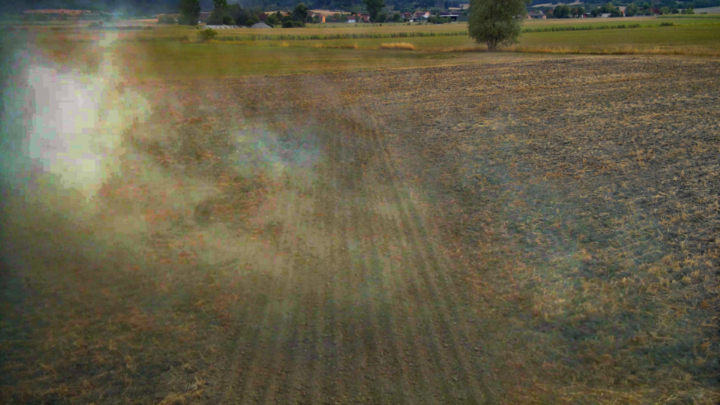}}\hfill
    \subfloat[CAP~\cite{zhu2015fast}]{\includegraphics[width=0.122\columnwidth]{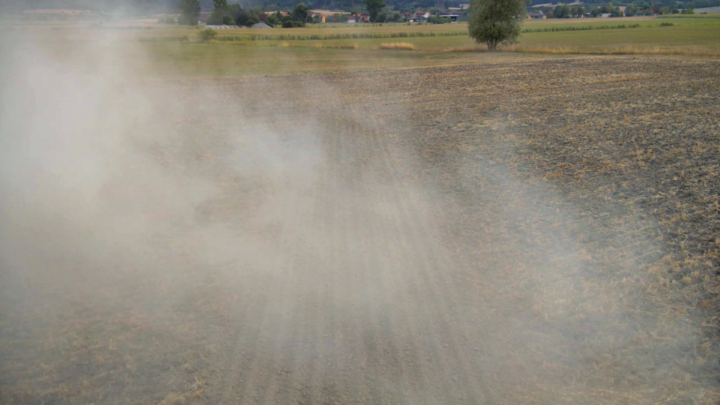}}\hfill
    \subfloat[CLAHE~\cite{zuiderveld1994contrast}]{\includegraphics[width=0.122\columnwidth]{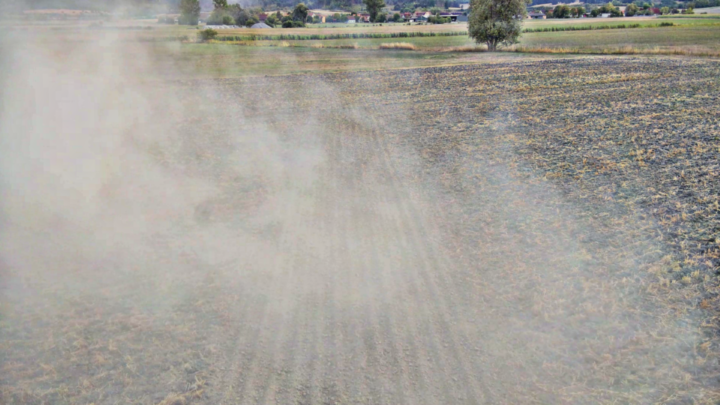}}\hfill
    \subfloat[Dust-free GT]{\includegraphics[width=0.122\columnwidth]{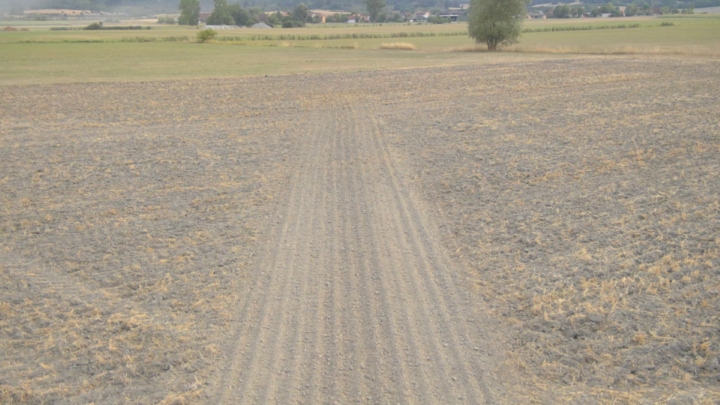}}
  \end{subfigure}
  \caption{Comparative results of different algorithms applied to the proposed dataset and the dust-free ground truth.}
  \label{fig:resultdata}
\end{figure*}
Image dehazing can be compared with image denoising. It does not distort the image truth but enhances it, thus improving its quality.
Overall, there are many different dehazing algorithms \cite{li2017aod, kim2021single, qu2019enhanced, dong2020fd, wang2022cycle, gao2014fast, li2022single, liu2022towards, yang2022self}. In the following we present the most common ones.
\newline
%CLAHE
A basic concept of increasing the contrast and therefore increasing noise reduction is histogram equalization (HE). HE stretches the pixel values over the entire range of values. Adaptive histogram equalization (AHE) always equalizes only the window of the image, and not the whole image~\cite{stark2000adaptive}. 
\newline
Contrast limited adaptive histogram equalization (CLAHE) limits the contrast by a specified value and equalizes histogram bins above~\cite{zuiderveld1994contrast}.
%DCP
\newline
He et al.~\cite{he2010single} developed dark channel prior (DCP), which is a prior for image dehazing. 
In the analysis of haze-free images, they showed that one color channel has a low value.
Based on their investigations, they have developed DCP, which shows the minimum intensity of each patch from the corresponding color channel.
Due to the similar characteristics of dusty and hazy images, DCP can also be used.
\newline
Zhu et al.~\cite{zhu2015fast} introduced a new prior called color attenuation prior (CAP). They investigated hazy images and used the difference between brightness and saturation. Based on this principle, they obtained a depth map by training a linear model using supervised learning. The scene radiance was recovered with the depth map and finally the haze-free image.
%NLD
\newline
In “Non-Local Image Dehazing” (NLD), Berman et al.~\cite{berman2016non} propose a new non-local instead of a patch-based prior. They discovered that the colors of a haze-free image could be described by clustered distinct colors in the RGB space. Within a cluster, these pixels are non-local. Due to the haze, these clusters change and build lines. Therefore, the authors developed an algorithm which detects these haze lines and estimates the transmission map. Consequently, their algorithm is faster. On the other hand, this method fails if the ambient light is brighter than the scene.
%DehazeNet
\newline
With DehazeNet, Cai et al.~\cite{cai2016dehazenet} proposed an end-to-end network for transmission estimation. 
The final haze-free image was obtained with the atmospheric scattering model. 
DehazeNet incorporates four steps for medium transmission estimation. 
First, it extracts the features with the help of the maxout unit as non-linear activation. In this way, it learns the features automatically. 
After that, it uses multi-scale mapping for multi-scale feature extraction and finds the local extremum to achieve spatial invariance. 
Finally, the authors propose a bilateral rectified linear unit (BReLU) for non-linear regression. The BReLU solves the problem of decreasing gradients of ReLU.
\newline
Ren et al.~\cite{ren2016single} developed a multi-scale convolutional neural network (MSCNN). Their work focused on the problem of selected features which come from prior-based dehazing methods, such as DCP~\cite{he2010single}. 
MSCNN was trained on synthesized data based on the NYU depth~\cite{Silberman:ECCV12} dataset. It consists of two steps. First, a coarse-scale network predicts a holistic transmission map of the entire image. Second, a fine-scale net uses previously generated results and adapts/refines these to local information. Therefore, the results contain global and local information. Ren et al. showed that their network outperforms state-of-the-art methods, and they were able to generalize the results to real-world images.
\subsection{Results}
In the next step, we applied the mentioned algorithms to the dataset. For comparability, we used the Matlab implementation of SSIM, PSNR, and NIQE.  
Table \ref{hazerestable} shows the results of the algorithms on the presented dataset. 
\begin{table}[h]
  \centering
  \resizebox{\columnwidth}{!}{%
  \centering
  \begin{tabular}{lcccccc}
  \hline
       & NLD~\cite{berman2016non}     & MSCNN~\cite{ren2016single}   & DehazeNet~\cite{cai2016dehazenet} & DCP~\cite{he2010single}     & CAP~\cite{zhu2015fast}             & CLAHE~\cite{zuiderveld1994contrast}            \\ \hline
  SSIM & 0.4791  & 0.7208  & 0.7407    & 0.7008  & \textbf{0.7671} & 0.6601           \\
  PSNR & 11.2350 & 16.3152 & 16.2885   & 14.2583 & \textbf{18.2541}         & 18.2335 \\
  NIQE & 3.2550  & 2.8973  & 2.8660    & 2.9252  & \textbf{2.8555} & 2.9327           \\ \hline
  \end{tabular}
  }
  \caption{Quantitative evaluation of state-of-the-art algorithms. The values are mean values of the entire dataset.}\label{hazerestable}
\end{table}
\newline
It demonstrates that NLD~\cite{berman2016non} does not offer good generalizability for dust. With NLD, lines are formed in the RGB space using distinct colors from haze. Because of this, the color selection is presumably not related to the dust. The two AI-based approaches MSCNN~\cite{ren2016single} and DehazeNet~\cite{cai2016dehazenet} give comparable results. 
CAP~\cite{zhu2015fast} outperforms other algorithms and reaches a SSIM of 0.7671, PSNR of 18.2541 and NIQE of 2.8555.
Figure \ref{fig:resultdata} shows the results of the algorithms. All the algorithms remove dust particles from images. Especially in low-density dust areas, the dust is completely removed. However, with increasing dust density the algorithms do not show any effect.
\subsection{Generalizability}
The dataset focuses on a static capture of images with and without dust. However, future autonomous agricultural machinery will have cameras mounted on or in the tractor. Images obtained in this way will then be evaluated for environmental perception.
\newline
Therefore, we took additional pictures to evaluate the performance of the methods. For this purpose, the camera was attached to the rear window in the tractor cabin. For the recording, the ground was again worked flat with the help of a disc harrow to maximize dust development. The main goal of generalization experiments is not to find the best method, but rather to verify the suitability of the test setup.
\begin{figure}[h]
   \centering
   \begin{subfigure}{0.99\linewidth}
    \subfloat[Dusty image during tillage]{\includegraphics[width=0.49\columnwidth]{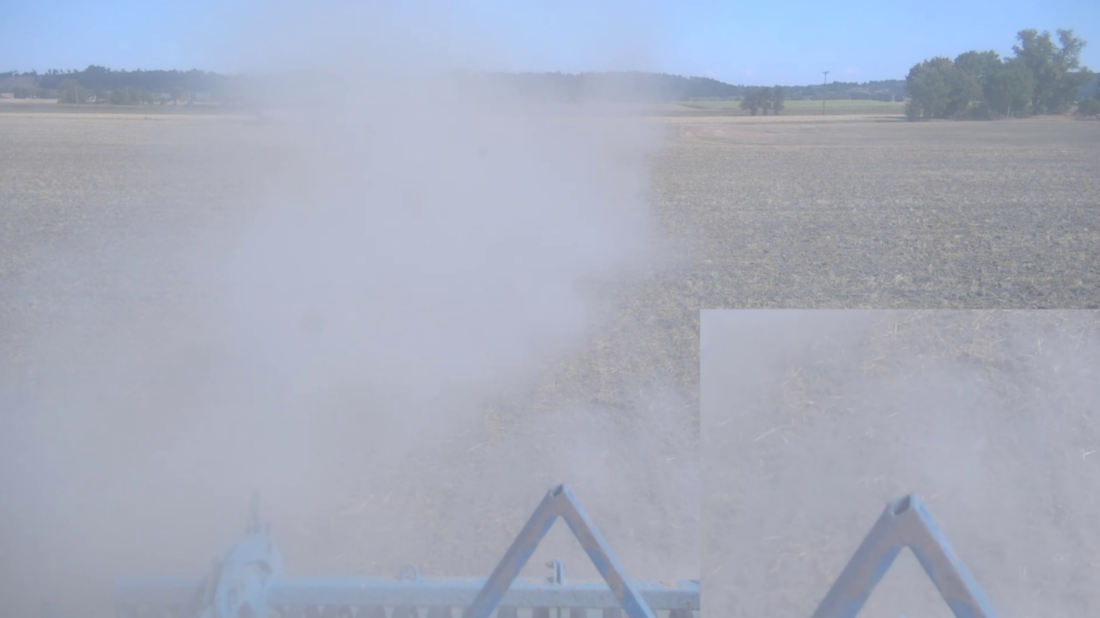}}\hfill
    \subfloat[Result of CAP~\cite{zhu2015fast}]{\includegraphics[width=0.49\columnwidth]{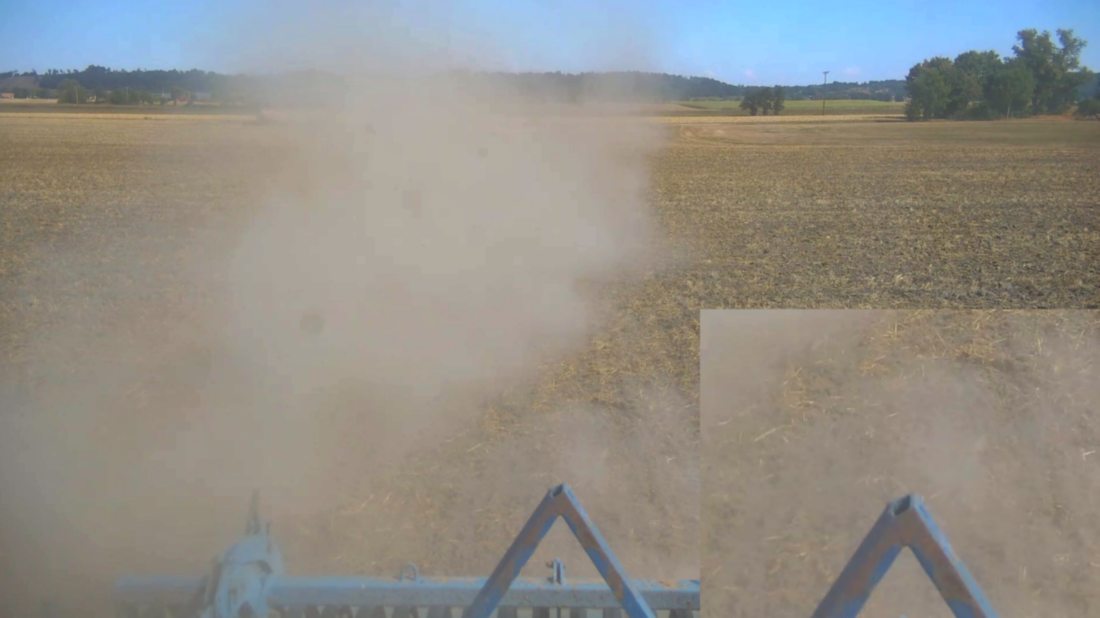}}\hfill
   \end{subfigure}
   \caption{Generalization to images out of the tractor cabin. A cropped part of the image can be seen in the bottom right corner.}
   \label{fig:resultdatagenera}
 \end{figure}
\newline
Based on the previous results we decided to use CAP~\cite{zhu2015fast} (Figure \ref{fig:resultdatagenera}). The lower right corner shows a section of the image which demonstrates that the method generalizes quite well to images taken out of the cabin. 
Thus, even as with images from the dataset, dust with a low-density is removed very well. Thereby, textures, edges, and objects are easier to recognize. However, the chosen algorithm reached its limits at high-density dust. Nevertheless, the results from the dataset can be transferred to images from the cabin.
\section{High-level Vision Task Experiments}
\begin{figure}[!h]
  \centering
  \includegraphics[width=0.99\columnwidth]{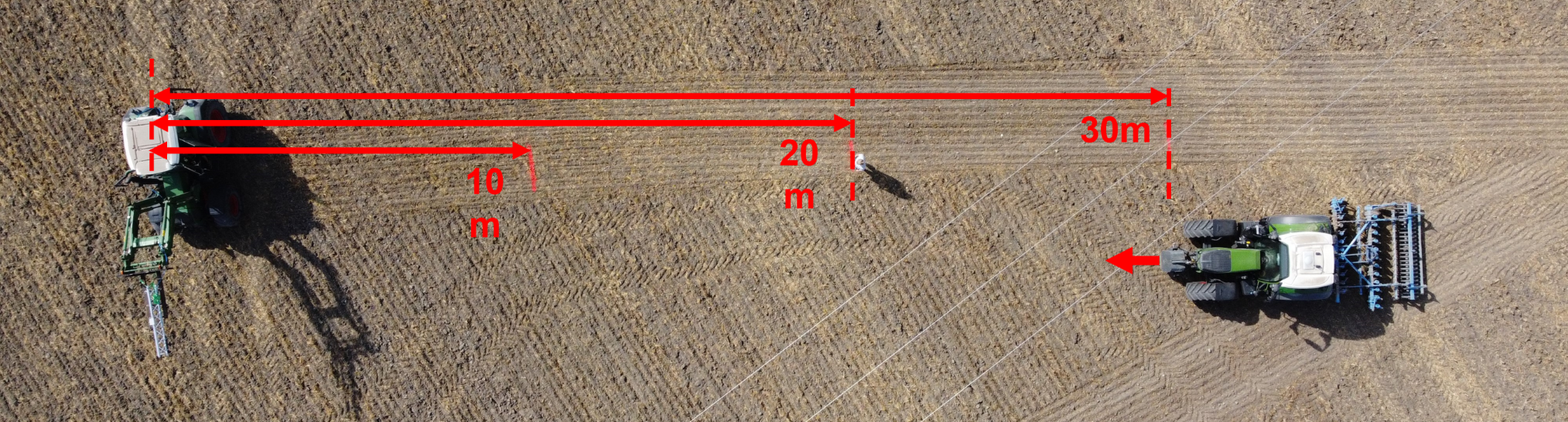}
  \caption{Test scenario for capturing images with a person standing still at 10 m, 20 m, and 30 m from the camera.}
  \label{fig:airperson}
\end{figure}

\begin{table*}[h]
  \centering
  \resizebox{0.9\linewidth}{!}{%
  \centering
  \begin{tabular}{lccccccccc}
  \hline
                      & \multicolumn{9}{c}{mAP}                                                        \\
                      & \multicolumn{3}{c}{10 m}  & \multicolumn{3}{c}{20 m}  & \multicolumn{3}{c}{30 m}  \\
                      & Light  & Medium & Heavy  & Light  & Medium & Heavy  & Light  & Medium & Heavy  \\ \hline
  Faster R-CNN        & 0.8683 & 0.8845 & 0.5000 & 0.6114 & 0.5033 & 0.0832 & 0.8203 & 0.5409 & 0.1040 \\
  CAP \& Faster R-CNN & 0.8931 & 0.8912 & 0.5160 & 0.6708 & 0.5168 & 0.2832 & 0.8340 & 0.6104 & 0.2218 \\
  Improvement         & 0.0248 & 0.0067 & 0.0160 & 0.0590 & 0.0135 & \textbf{0.2000} & 0.0137 & 0.0695 & \textbf{0.1178} \\ \hline
  \end{tabular}
  }
  \caption{Comparison of mAP with / without CAP~\cite{zhu2015fast} and the improvement at 10 m, 20 m, and 30 m.}\label{personrestable}
\end{table*}
Next, we contemplated how algorithms could be applied in agriculture in the future. The most common use cases of images are classification and recognition tasks. 
Object detection and recognition is already an extensively researched field. 
In the work of \cite{pei2019effects, li2017aod, li2018end, pei2020consistency}, for example, the possible impact of dehazing algorithms on classification tasks is already being investigated and discussed.
Overall, researchers come to different conclusions, and therefore we added more images with people to our dataset.
We are not aware of any available open-source datasets that contain persons who are covered by dust. 
Figure \ref{fig:airperson} shows the testing scenario.
\newline
\newline
A person stood as close as possible to the tractor track at defined distances from the camera and thus from the gate. We then drove past the person with the second tractor. 
The whirled-up dust covered the person completely or partially. Moreover, the defined distances of 10 m, 20 m, and 30 m allowed us to identify the detection probability of a classification network in relation to the distance. In addition, further investigations can be made to determine to what extent algorithms for dust removal improve the classification rate.
The dataset for person detection includes 45 images from one background. These cover the whole range of different dust levels: from no dust to very dense dust.
\newline
First, we clustered the dusty images into different dust levels. We divided them into light, medium, and heavy dust and selected five images for each level and distance. The images of the different levels were selected based on the following criteria:
\newline
\newline
\textbf{Light dust:}
\begin{itemize}
  \item Person is fully visible and covered by minor dust.
\end{itemize}
\textbf{Medium dust:}
\begin{itemize}
  \item Person is partially covered by non-translucent dust.
  \item Person is fully covered but still visible.
\end{itemize}
\textbf{Heavy dust:}
\begin{itemize}
  \item Person is almost completely or totally covered with heavy dust that is not transparent.
\end{itemize}
Afterwards, the person still has to be detected. For our model, we chose a Faster R-CNN~\cite{ren2015faster}, since its accuracy and the combination with dehazing algorithms has already been extensively demonstrated in various works~\cite{li2017aod}. 
Therefore, we chose a pre-trained Faster R-CNN, which is improved with a ResNet backbone~\cite{li2021benchmarking}. The model is available at~\cite{fasterRCNNpytorch}. The network is pre-trained on the COCO dataset, which includes over 200,000 labeled images and 80 object classes~\cite{lin2014microsoft}.
We applied the Faster R-CNN~\cite{li2021benchmarking} to the dusty and enhanced images and calculated the mean average precision (mAP), using a threshold of 0.5. We chose CAP~\cite{zhu2015fast} as the image enhancement algorithm based on the previous experiments.
\newline
Table \ref{personrestable} shows the results of the experiments and the difference between the two, i.e., the improvement.
At 10 m, the mAP was only slightly higher. In contrast, at 20 m, mAP increased by 5.9\% for light dust and by as much as 20\% for heavy dust.
There was a 6.95\% increase in mAP for medium dust at 30 m and 11.78\% for heavy dust.
\begin{figure*}[h]
   \centering
   \begin{subfigure}{0.99\linewidth}
     \includegraphics[width=0.245\columnwidth]{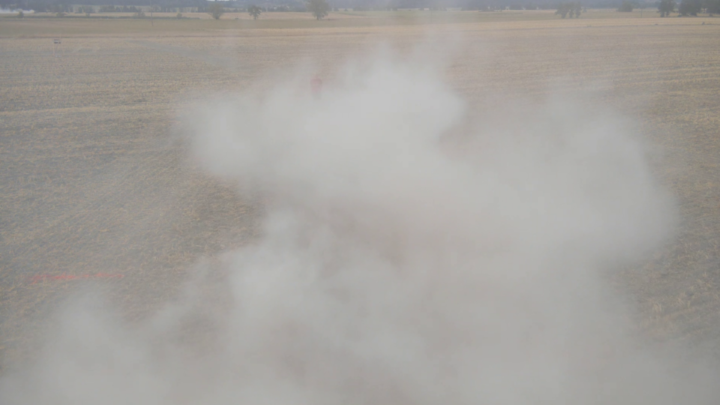}\hspace{.01mm}
     \includegraphics[width=0.245\columnwidth]{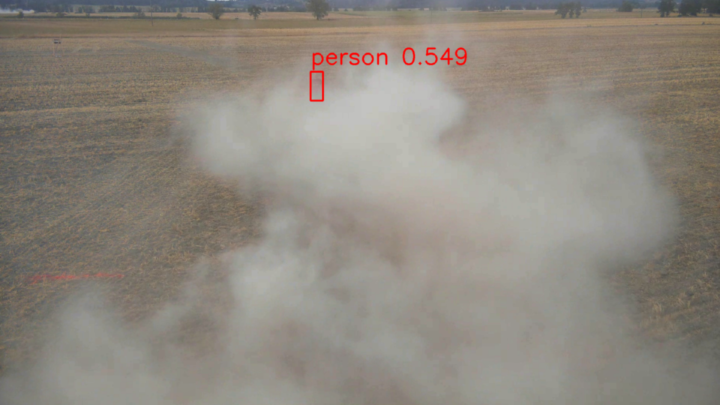}\hfill
     \includegraphics[width=0.245\columnwidth]{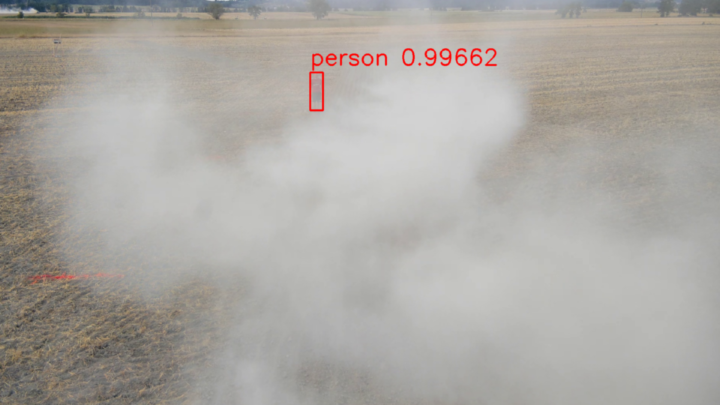}\hspace{.01mm}
     \includegraphics[width=0.245\columnwidth]{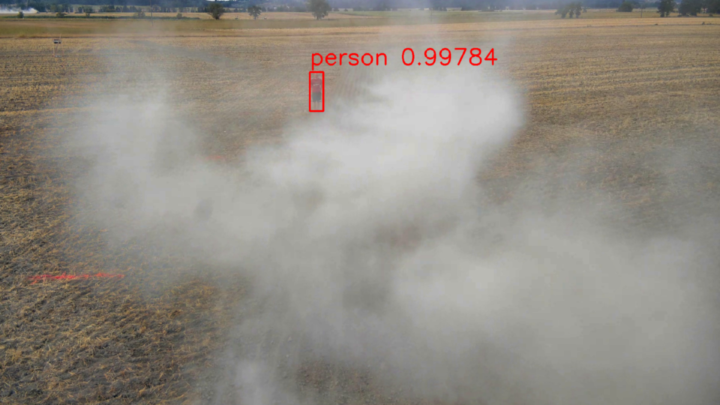}\hfill
     \qquad
     \includegraphics[width=0.245\columnwidth]{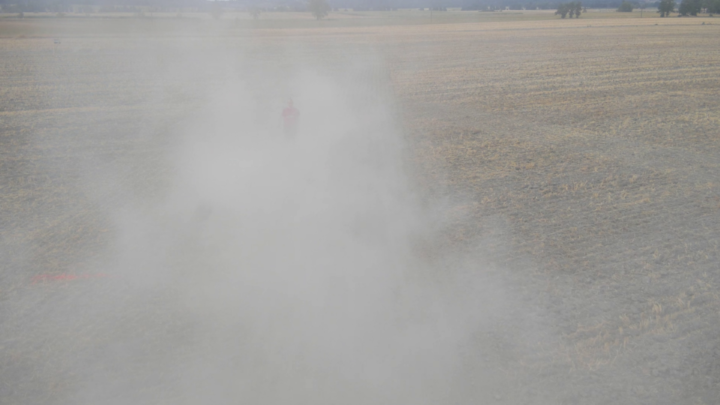}\hspace{.01mm}
     \includegraphics[width=0.245\columnwidth]{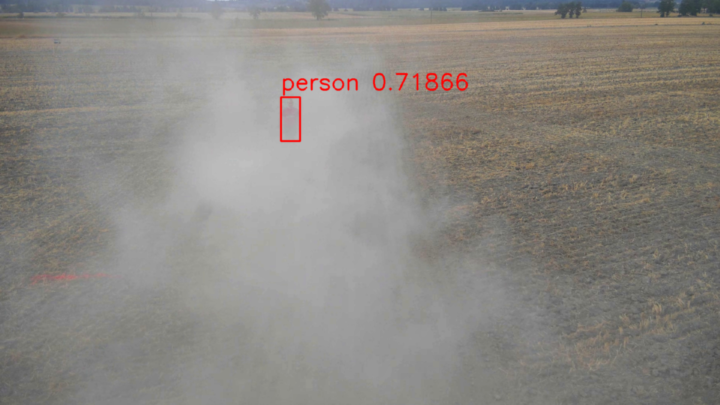}\hfill
     \includegraphics[width=0.245\columnwidth]{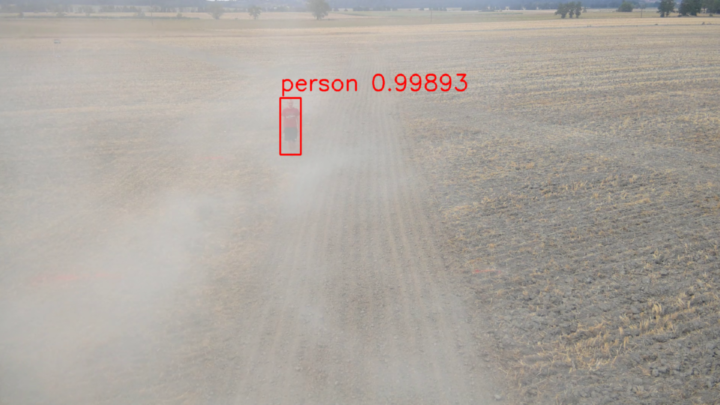}\hspace{.01mm}
     \includegraphics[width=0.245\columnwidth]{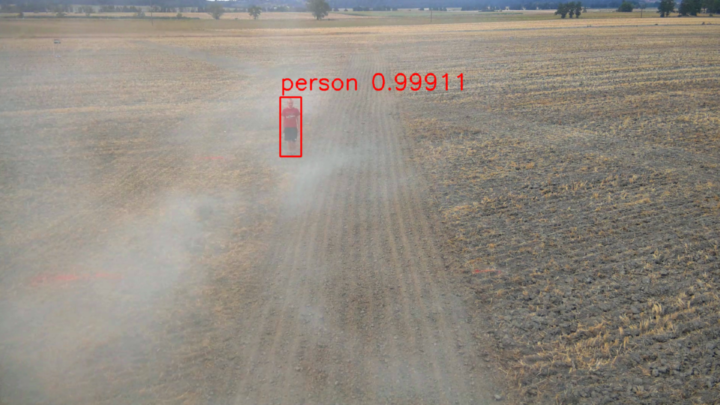}\hfill
     \qquad
     \subfloat[Faster R-CNN]{\includegraphics[width=0.245\columnwidth]{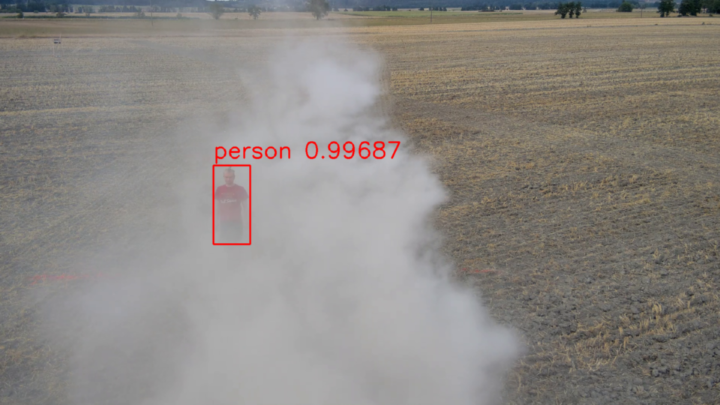}}\hspace{.01mm}
     \subfloat[CAP \& Faster R-CNN]{\includegraphics[width=0.245\columnwidth]{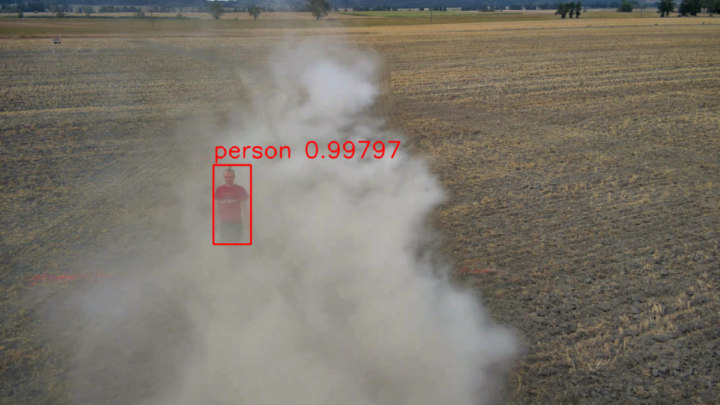}}\hfill
     \subfloat[Faster R-CNN]{\includegraphics[width=0.245\columnwidth]{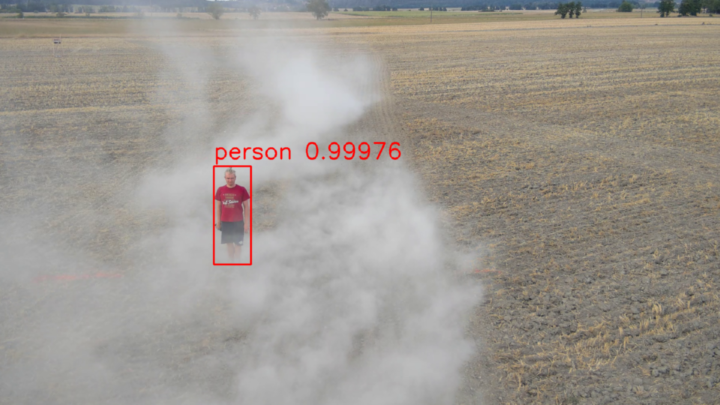}}\hspace{.01mm}
     \subfloat[CAP \& Faster R-CNN]{\includegraphics[width=0.245\columnwidth]{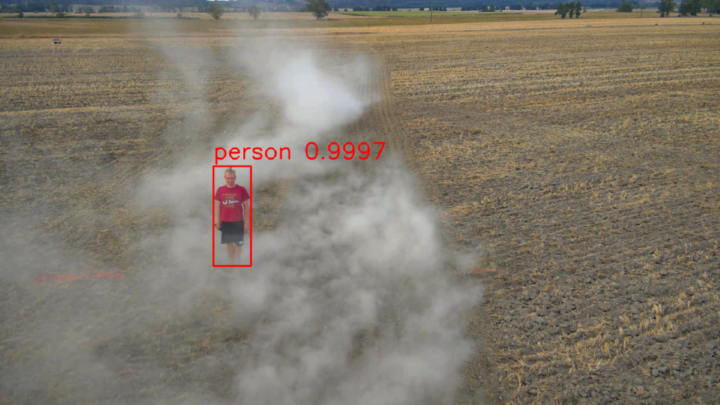}}\hfill
   \end{subfigure}
   \caption{Person detection and recognition results with / without CAP~\cite{zhu2015fast} applied to different levels of dust. First row 30 m (heavy and medium), second 20 m (heavy and light), and third row 10 m (heavy and light).}
   \label{fig:resultpersongatedist}
 \end{figure*}
\newline
Figure \ref{fig:resultpersongatedist} shows the results of the pre-trained Faster R-CNN with CAP:
At short distances Faster R-CNN recognizes people both without and with the applied algorithm. No difference can be seen with the naked eye.
At 20 m and 30 m, especially in heavy dust, the Faster R-CNN fails to detect the person. In contrast, after enhancement, the person could be spotted. 
Furthermore, CAP leads to better detection of the person at these distances in medium and light dust. For example, parts of the feet can be seen.
\section{Discussion}
Our reference-based dataset for dust removal has some limitations. First, the proposed gate, and therefore the proposed dataset, is not noise-free. 
Vibrations of the gate and noise in the air cause minor deviations. 
This means that even a perfect algorithm would not achieve the identical dust-free reference. 
For the evaluation of the proposed RB-Dust dataset, we used algorithms from the field of dehazing and were able to show that CAP~\cite{zhu2015fast} gives the best results in terms of SSIM and NIQE. Presumably, algorithms designed to remove dust instead of haze are better.
Furthermore, it is important that due to the higher position of the camera in the images, the proportion of the sky in the images is lower than from possible future positions in the cabin. This, in turn, affects the generalizability of the results, but we were able to show that it is still possible to do.
For person detection, we used a pretrained network and were able to show that removing dust increases the mAP by up to 20.00\%. 
However, the database was limited to five images per dust density level and distance. 
Furthermore, it must be noted that classification networks trained on images with people in the dust may be better detecting them.
This could mitigate the effect of dust removal on high-level vision tasks.
Our results encourage further investigations in the area of dust removal.
\section{Conclusion}
In this paper, we proposed the agriscapes RB-Dust dataset for reference-based dust removal in agriculture and presented our measurement setup, which is based on a boom mounted to the front loader of a tractor. The proposed gate allowed us to record images from same scene with and without dust. In addition, our final dataset consisted of 200 images taken from different positions and at different times, thus covering a variety of angles and light conditions. Furthermore, we were able to prove that the setup is only subject to very small fluctuations and is therefore suitable for further investigations. The dataset was validated with prior- and CNN-based dehazing methods. It was possible to remove dust partially and generalize the results into the tractor cabin. Finally, we investigated the effect of dust removal to high-level vision tasks such as person detection. Based on our preliminary results, the dust removal algorithms can improve person recognition.
\newpage

%%%%%%%%% REFERENCES
{\small
\bibliographystyle{ieee_fullname}
\bibliography{egbib}
}

\end{document}